\documentclass[lettersize,journal]{IEEEtran}
\pdfoutput=1
\usepackage{cite}
\usepackage{amsmath,amssymb,amsfonts}
\usepackage{graphicx}
\usepackage{textcomp}

\usepackage{fancyhdr}
\usepackage{xspace}
\usepackage{pifont}
\usepackage{booktabs}
\usepackage{multicol}
\usepackage{subcaption}

\usepackage{comment}
\usepackage{pdfpages}
\usepackage{hyperref}
\usepackage{latexsym}
\usepackage{multirow}
\usepackage{balance}

\usepackage{tabulary}
\usepackage{bm}
\usepackage{diagbox}

\usepackage[ruled,vlined]{algorithm2e}
\SetKwInput{KwInput}{Input}
\SetKwInput{KwOutput}{Output}
\SetKwRepeat{Do}{do}{while}

\newcommand{\bb}{\color{black}}

\newcommand{\ie}{\textit{i.e.}\xspace}

\newcommand{\eg}{\textit{e.g.}\xspace}

\newcommand{\etc}{\textit{etc.}\xspace}

\definecolor{realdataColorCode}{RGB}{30, 136, 229}
\definecolor{augmentColorCode}{RGB}{0,150,136}

\definecolor{light-gray}{gray}{0.7}
\newcommand{\cmark}{\text{\ding{51}}}%
\newcommand{\xmark}{\color{light-gray} \text{\ding{55}}}%

\newcolumntype{L}[1]{>{\raggedright\let\newline\\\arraybackslash\hspace{0pt}}m{#1}}
\newcolumntype{C}[1]{>{\centering\let\newline\\\arraybackslash\hspace{0pt}}m{#1}}

\usepackage{tabularx}

\newcommand\clearrow{\global\let\rowmac\relax}

\newcolumntype{R}[2]{%
    >{\adjustbox{angle=#1,lap=\width-(#2)}\bgroup}%
    c%
    <{\egroup}%
}
\begin{document}

\title{LatentAugment: Data Augmentation via Guided Manipulation of GAN's Latent Space}

\author{Lorenzo Tronchin, Minh H. Vu, Paolo Soda, and Tommy L\"{o}fstedt
\thanks{Lorenzo Tronchin is with the Department of Engineering, University Campus-Biomedico of Rome, Rome, Italy (e-mail: l.tronchin@unicampus.it).}
\thanks{Minh H. Vu, is with the Department of Radiation Sciences, Ume{\aa} University, Ume{\aa}, Sweden (e-mail: minh.vu@umu.se).}
\thanks{Paolo Soda is with the Department of Engineering, University Campus-Biomedico of Rome, Rome, Italy (e-mail: p.soda@unicampus.it) and with the  Department of Radiation Sciences, Ume{\aa} University, Ume{\aa}, Sweden (e-mail: paolo.soda@umu.se).
}
\thanks{Tommy L\"{o}fstedt is with the Department of Computing Science, Ume{\aa} University, Ume{\aa}, Sweden (e-mail: tommy.lofstedt@umu.se).}
}

\maketitle

\begin{abstract}
Data Augmentation (DA) is a technique to increase the quantity and diversity of the training data, and by that alleviate overfitting and improve generalisation.
However, standard DA produces synthetic data for augmentation with limited diversity.
Generative Adversarial Networks (GANs) may unlock additional information in a dataset by generating synthetic samples having the appearance of real images.
However, these models struggle to simultaneously address three key requirements: fidelity and high-quality samples; diversity and mode coverage; and fast sampling.
Indeed, GANs generate high-quality samples rapidly, but have poor mode coverage, limiting their adoption in DA applications.
We propose LatentAugment, a DA strategy that overcomes the low diversity of GANs, opening up for use in DA applications.
Without external supervision, LatentAugment modifies latent vectors and moves them into latent space regions to maximise  the synthetic images' diversity and fidelity.
It is also agnostic to the dataset and the downstream task.
A wide set of experiments shows that LatentAugment improves the generalisation of a deep model translating from MRI-to-CT beating both standard DA as well GAN-based sampling. 
Moreover, still in comparison with GAN-based sampling, LatentAugment synthetic samples show superior mode coverage and diversity.
Code is available at: \href{https://github.com/ltronchin/LatentAugment}{https://github.com/ltronchin/LatentAugment}.
\end{abstract}


\section{Introduction} \label{sec:introduction}

Deep learning has recently had many successes in decision tasks, especially when large amounts of data are available~\cite{deng2009imagenet}.
Several explicit or implicit regularisation techniques have been developed to overcome overfitting in cases with less data, such as dropout~\cite{hinton2012improving}, batch normalisation~\cite{ioffe2015batch}, or transfer learning~\cite{zhuang2020comprehensive}.
However, these methods cannot exploit known input invariances that form constraints for parameter learning, especially not in the regimes with small amounts of data~\cite{antoniou2017data}.

To cope with this issue, Data Augmentation (DA) has been widely utilised to improve generalisation and robustness when training deep neural networks~\cite{krizhevsky2017imagenet}. 
Common DA methods in image recognition tasks transform the images via geometric rigid and non-rigid transformations, using image processing primitives, such as \eg, translation, rotation, cropping, \etc~\cite{chlap2021review}.
However, in most cases, designing such transformations has relied on human experts with prior knowledge of the dataset.
Indeed, even if useful augmentations have been found for a given dataset, they may not transfer to other datasets.
For example, horizontal flipping of images during training is an effective data augmentation method for CIFAR-10 (natural images) but not for MNIST (hand-written digits) due to the different symmetries present in these datasets~\cite{cubuk2019autoaugment}.

Recently, some efforts have been directed towards designing an automated process to search for augmentation policies directly from a target dataset~\cite{cubuk2019autoaugment, tang2020onlineaugment, cubuk2020randaugment, liu2021divaug}.
Unfortunately, they still provide restricted variations in the data, limiting the invariances that a target model can learn~\cite{antoniou2017data}.
Indeed, they rely on pre-specified image processing functions as augmentation operations. 
Defining the basic operations requires domain knowledge, which may impede their application in more tasks.
Moreover, these approaches often use reinforcement learning and their application requires thousands of GPU hours.

Generative Adversarial Networks (GANs) offer a valuable addition to the available set of augmentation techniques.
A GAN learns to generate samples from the distribution of training samples, thereby considering all sources of variation within the data.
For example, given sufficient training examples of patients with different ventricle sizes, a GAN will learn to generate samples along the continuum of all ventricle sizes.
Thus, manipulating the samples generated by GANs allows very complex image transformations.
Vahdat~\textit{et~al.}~\cite{xiao2021tackling} stated three key requirements for generative frameworks to be adopted for real-world problems, including: (i) fidelity, meaning the quality of the generated samples, particularly their realism; (ii) diversity and mode coverage, meaning the variation and variety of the samples that can be generated; and (iii) how fast samples can be generated.
The authors identify the challenge posed by these requirements as the ``the generative learning tri\-lemma'' and concluded that generative models compromise between them.
GANs generate high-quality samples rapidly, but they suffer from poor mode coverage.
Our hypothesis is that the lack of diversity of GAN-generated images can hinder their applicability for DA purposes.
Indeed, existing generative methods generate purely random images without control over what images are generated.

On these grounds, here we address the generative learning trilemma for GANs, and improve their effectiveness for DA purposes by
proposing LatentAugment, a new GAN-based augmentation policy that maximises the diversity, \ie, the variability in the generated data with respect to the training data distribution; and that maximises the fidelity, \ie, how similar the generated images are to real images (high-quality generation). 
The proposed policy works in the GAN latent space, which reduces the computational cost compared to working in the image space, and exploits the semantic information that the generator has learned.
The rest of the manuscript is organised as follows:
\sectionautorefname~\ref{sec:related_work} introduces the state-of-the-art of DA methods and the motivation of LatentAugment. 
Then, \sectionautorefname~\ref{sec:methods} presents our novel DA method.
In \sectionautorefname~\ref{sec:experiments} we describe the dataset used to validate the method, the pre-processing phase on the data, the GAN architecture we use, other DA approaches tested for comparative analysis, and the validation strategy adopted.
Section \ref{sec:results} presents and discuses the obtained results, whilst \sectionautorefname~\ref{sec:conclusion} provides concluding remarks.

\section{Background and motivations} \label{sec:related_work}

It is challenging to obtain reliable generalisation in practical applications with small datasets, for example in medical imaging where it remains expensive to acquire informative and noise-free annotations~\cite{litjens2017survey}.
DA increases the size of the training set by artificially creating new samples and reduces the risk of overfitting when training DL models on datasets of limited sizes~\cite{wong2016understanding,taylor2018improving}.
In the rest of this section we adopt the taxonomy on image augmentation techniques proposed by Xu \textit{et al.}~\cite{xu2023comprehensive}. 
They divided DA methods into three main branches: \textit{model-free}, \textit{optimising policy-based}, and \textit{model-based}. 

\subsection{Model-free}
\label{subsec:model_free}

Model-free techniques leverage image processing methods, such as geometric transformations and pixel-level manipulation, and are further divided into single-image augmentation and multiple-image augmentation.

Well-know single-image augmentation approaches on natural images include horizontal flips, random cropping, rotation, and translations.
These techniques have, for instance, been used in classification and detection tasks~\cite{krizhevsky2017imagenet,he2017mask}.
Such approaches simulate intra-class variation, augmenting the data while keeping them close to the training set, \ie, they explicitly teach the model to be invariant to the particular transformations used.
Other single-image approaches that vary the data more and increase generalisation further. 
Within this category, intensity transformation changes the image at pixel or patch level: for instance, the former could add independent random noise to be robust to artefacts in the image generation~\cite{vapnik1991principles}, whereas the latter could achieve invariance to occlusions~\cite{singh2017hide,devries2017improved,zhong2020random,chen2020gridmask}. 

Multiple-image augmentation methods are executed on more than one image and aim to merge multiple inputs~\cite{inoue2018data,zhang2017mixup,yun2019cutmix}.
Examples include SamplePairing~\cite{inoue2018data} and Mixup~\cite{zhang2017mixup}.
In SamplePairing, the images are averaged, and a label is selected among the source images. In Mixup, models are trained on a convex combination of the images and their labels.

Unfortunately, both single- and multiple-image augmentation methods require expertise and manual work to design policies  
tailored to the domain at hand.
This, in turn, requires hyper-parameter optimisation of the transformation settings, such as the probability or the magnitude of the augmentation that is applied, making it difficult to apply a DA policy from one domain to another one.

\subsection{Optimising policy-based}

Learning policies for data augmentation have emerged as an approach to automate augmentation strategies to overcome the weaknesses of model-free methods.
These approaches aim to select a well-suited set of augmentation functions, \eg, rotation, shift, \etc, for the dataset at hand 
and use reinforcement or adversarial learning~\cite{cubuk2020randaugment,cubuk2019autoaugment,lim2019fast,ho2019population,zhang2019adversarial,tang2020onlineaugment}.

For instance, AutoAugment~\cite{cubuk2019autoaugment} finds the best set of transformations for a proxy task.
Extensions and improvements have been proposed~\cite{cubuk2020randaugment,lim2019fast,ho2019population}.

Adversarial training improves the robustness of downstream models by augmenting with difficult samples~\cite{zhang2019adversarial,tang2020onlineaugment}, \ie, samples that cause a high training loss.
The assumption is that they are useful to improve the generalisation of deep models.

Optimising policy-based strategies learn the augmentation method, improving a downstream task. Nevertheless, they are limited to a set of known, pre-defined transformations, constraining the invariances that are introduced to the dataset.

\subsection{Model-based} \label{subsec:ModelBased}

Model-based augmentation simultaneously modifies the style and content of the images, aiming to extend the possible created variance while maintaining fidelity.
Such methods use synthetic images a generative model produces to enlarge the original dataset~\cite{xu2023comprehensive}.
GANs have attracted increased attention due to their remarkable image-generation performance and have been used for segmentation and classification~\cite{Frid-Adar_etal_2018b,calimeri2017biomedical,madani2018chest,sandfort2019data}.
GANs provide a way to augment sources of variance in the data that would be challenging to define otherwise. For instance, a GAN trained on a set of images of an organ can synthesise images where the organ varies its size with continuity~\cite{liu2020manipulating}.
Thus, they are able to introduce a type of variance not straightforward to capture with common augmentation methods~\cite{xu2023comprehensive}.
Skandarani \textit{et~al.}~\cite{skandarani2023gans} argued that GANs could not reproduce the full richness of medical datasets, motivated by experimental results in segmentation where no deep networks trained on large numbers of real and synthetic samples outperform networks trained on only real data.

State-of-the-art GAN augmentation techniques randomly generate synthetic images~\cite{yi2019generative,chen2022generative,skandarani2023gans}, but have no control over the diversity in the generated images.
This is in contrast to the original idea of DA that assumes that effective data transforms should produce samples from an ``overlapping but different'' distribution~\cite{bengio2011deep,bellegarda1992robust}.
Thus, the lack of control over the GAN-generated images \textit{de facto} limits their efficacy for DA and is identified as one of the main reasons for the poor performance observed when using synthetic data~\cite{chen2022generative}.

\subsection{Motivations} \label{subsec:motivations}

DA has been affirmed as a technique to artificially increase training set sample variability by transforming data points in a way that, in supervised learning, preserves class labels, and has become an effective tool for tackling data scarcity problems~\cite{ratner2017learning}.
However, the choice of DA strategy is known to cause large variations in downstream performance and can be difficult to select~\cite{cubuk2019autoaugment}.
While recent works based on optimising policies have attempted to automate DA~\cite{xu2023comprehensive}, they only consider restricted sets of simple transformations, thus limiting the invariances a downstream model can learn.
GANs have the potential to take many decisions away from the user, in much the same way as deep learning removed the need for hand-crafted features~\cite{bowles2018gan}.
However, the main drawback of GAN-based augmentation remains the lack of control over the generated images.
While GANs generate high-quality samples rapidly, and thus fulfilling the first two criteria of the generative learning trilemma, they suffer from poor mode coverage, limiting their effectiveness for DA application.

To address this limitation, we propose a novel GAN-based augmentation method that controls the generation of synthetic images in the latent space to improve diversity and fidelity.
It is worth noting that the idea of using the latent space has its roots in image editing techniques, which have tackled the issue of lack of direct control over the GAN generation.
Indeed, they aim to learn how to navigate the latent space in directions that allow changing the semantics of generated images~\cite{xia2022gan}, such as facial attributes~\cite{shen2020interfacegan}, memorability of images~\cite{goetschalckx2019ganalyze}, or camera movements and colour changes~\cite{jahanian2019steerability}.
As far as we can tell, this work is not only the first attempt to use latent space manipulation in the context of data augmentation but also the first applied  to medical imaging.

\section{Methods} \label{sec:methods}
We propose an augmentation policy, referred to as \textit{Latent\-Augment}, that creates a new sample by navigating the latent space of a trained GAN to maximise the diversity of the augmented images while ensuring their fidelity.
In this section, we first formulate the DA problem and then detail how LatentAugment works.

\subsection{Problem Formulation} \label{subsec:problemFormulation}
Let ${\mathcal{X}}=[x_{1}, x_{2}, ..., x_{N}]$ denote a training set that consists of $N$ images.
The primary objective of a DA procedure, $\mathcal{A}$, is to train a downstream model $\mathcal{M}$ on augmented versions of the images, such that the downstream model generalises better to an independent test set, or to other new data.
During training, a common approach is to take a data point, $x_{i}$, and, before presenting it to $\mathcal{M}$, compute an augmented image $\tilde{x}_{i}$, as
\begin{equation}
  \tilde{x}_{i}=\begin{cases}
      \mathcal{A}(x_{i}) & \text{if $r \geq p_{aug}$},\\
      x_{i}              & \text{otherwise},
  \end{cases} \label{eq:AugmentRule}
\end{equation}
where $r$ is a random uniform real number in $[0, 1)$ and $p_{aug}\in[0,1]$ is a threshold probability whether to apply the DA procedure at all (activating or deactivating the DA procedure for that image).
We denote the augmented training set as $\widetilde{\mathcal{X}}$.
When $p_{aug}=0$, the downstream model $\mathcal{M}$ is fed only augmented images, and with $p_{aug}=1$, the augmentation procedure is disabled.
This approach thus does not increase the cardinality of the training dataset, but instead adds a layer of stochasticity to the learning process of  $\mathcal{M}$, and (ideally, but depending on $\mathcal{A}$) increases the diversity of the training data.

\subsection{Overview of the framework}

Here we introduce the two main ingredients of the proposed method: GANs and GAN-inversion.

A GAN consists of two networks, a generator $G$ and a discriminator $D$.
Inspired by game theory, those two networks are trained in an adversarial process where $G$ generates fake images attempting to fool the discriminator to believe that they are real, while $D$ attempts to discriminate between the real and fake images~\cite{goodfellow2020generative}.
The training process can be described as a min-max game,
\begin{align}
    \min_{G} \max_{D} V(G, D) 
        & = 
            \mathbb{E}_{x \sim \mathcal{X}}
            \big[\log D(x)\big]
            \nonumber \\ 
        & \;\qquad + 
            \mathbb{E}_{z \sim \mathcal{Z}}\Big[\log \Big(1 - D\big(G(z)\big)\Big)\Big],
            \label{eq:minimaxgame-bsg}
\end{align}
where the optimisation is over the parameters of $G$ and $D$, $\mathcal{X}$ is 
the data used to train the GAN, $z$ is the noise vector sampled from the latent space, $\mathcal{Z}$.
While generating new images, the generator takes a latent vector, $z$, and maps it to an image.
The min-max loss in \equationautorefname~\ref{eq:minimaxgame-bsg} and the training process guarantee that the estimated image manifold is aligned with the training image manifold.

In this work, we used the StyleGAN2 (SG2) architecture as the GAN backbone since it is the state-of-the-art GAN model for high-resolution image synthesis~\cite{karras2020analyzing}.
Unlike a traditional generator, the SG2 model introduces a multi-layer perceptron, $\mathcal{F}$, that maps $z$ to an intermediate latent space, $\mathcal{F}(z) = w \in \mathcal{W}$.
The generator, $G$, then synthesises images based on these intermediate latent vectors, $w$.
The latent space mapping allows the generator to learn an intermediate latent space, $\mathcal{W}$, that is less entangled by design~\cite{karras2019style,karras2020analyzing}, \ie, each dimension of $w$ controls only a single (or a few) features of the generated image.
A disentangled feature space is a key desiderata for any GAN-based image editing technique and, hence, also for the LatentAugment method proposed here.

GANs lack the ability to find the latent representation of an input image, which is a necessary step to manipulate images in the latent space for DA purposes.
Thus, we exploit GAN-inversion to reverse (invert) the mapping of $G$, to find a latent vector that recovers a given input image~\cite{xia2022gan}. With the SG2, this is an intermediate latent vector, $w^{*}$, but when using a GAN without a mapping network such as $\mathcal{F}$~\cite{radford2015unsupervised}, the inversion instead seeks a latent vector $z^{*}\in\mathcal{Z}$.

Existing inversion approaches are either learning- or optimisation-based.
The former involves training an encoding network to map an input image into the latent space, such that the found latent vector reproduces the input image (directly learning the inverse mapping).
The latter directly optimises in the latent space, searching for a latent vector that would regenerate the real input image (not explicitly learning an inverse mapping).
The first approach provides a fast solution for image embedding by performing a forward pass through the encoder, but does not generalise beyond the training dataset since it needs to be trained for each inversion task~\cite{abdal2019image2stylegan}.
We, therefore, adopted the second approach, using the optimisation-based method proposed by Karras \textit{et al.}~\cite{karras2020analyzing}, which is well-suited for inverting real images in the SG2 latent space, $\mathcal{W}$.

\begin{figure*}[!h]
    \centering
    \includegraphics[width=\textwidth]{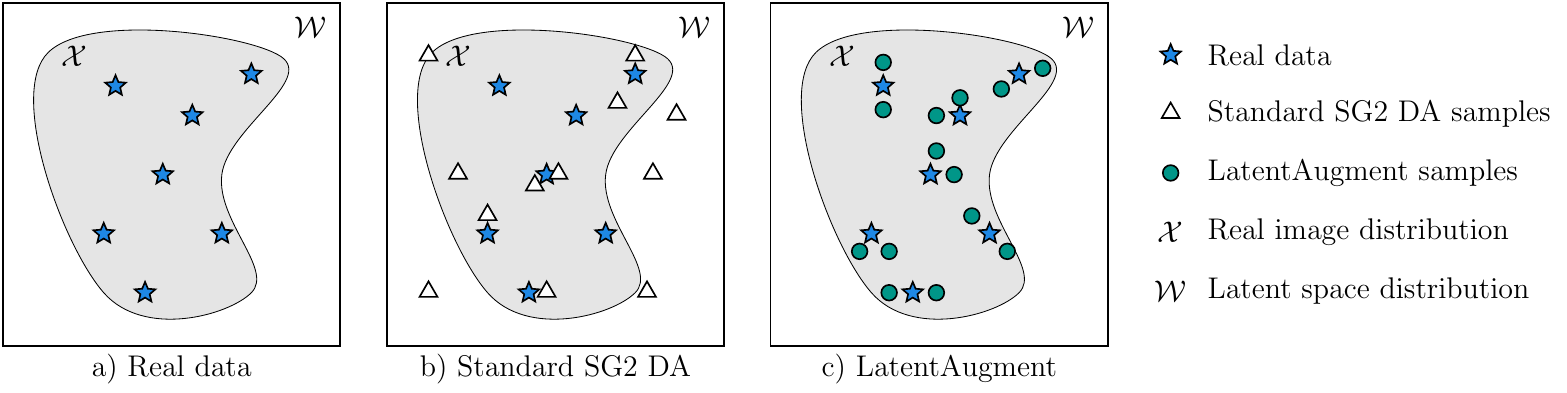}
    \caption{Intuition. Considering $\mathcal{X}$ and $\mathcal{W}$ as the data and latent space distributions, respectively. In panel (a), we denote with blue stars the encoded real images.
    Panel (b). The standard SG2 DA procedure does not provide any guide to the generation process. Thus random latent samples (white triangles) may overlap the manifold of the latent representations of the real images but may also fall arbitrarily far outside the real image distribution, $\mathcal{X}$.
    Panel (c). The LatentAugment method adds control to the GAN generation process allowing the synthetic latent vectors to be close to the latent vectors of the real images but not too close (green circles).}
    \label{fig:intuition}
\end{figure*}

\subsection{LatentAugment Policy}

\subsubsection{Intuition}

GANs are commonly used to synthesise new images or other data, but current methods do not allow control over the generation process, especially not so for DA.
To augment the training set of a downstream model, SG2-based policies compute $\Tilde{x} = G(w)$, where $w = \mathcal{F}(z)$ with $z$ randomly sampled from $\mathcal{Z}$.
This approach, referred to as \textit{Standard SG2 DA} in the following, is illustrated in panels a) and b) of \figureautorefname~\ref{fig:intuition}.
In this figure, the blue stars illustrate the latent positions of the real samples (retrieved through the inversion procedure), and the white triangles denote randomly sampled points in the latent space that will generate the synthetic images for a Standard SG2 DA procedure.
The Standard SG2 DA does not guarantee that such generated images are useful for the downstream task since they may lie outside of the manifold of the real data, illustrated by the shaded area in \figureautorefname~\ref{fig:intuition}, and may correspond to cases where generated images contain artefacts or are of low quality---such synthetic images would thus have low fidelity.
If the generator, $G$, overfits the training data, the synthetic images would look much like the training images themselves, \ie, in the white triangles would overlap the blue stars in \figureautorefname~\ref{fig:intuition}. Such generated images have low diversity but high fidelity~\cite{tronchin2021evaluating}.

To formulate a GAN-based DA policy, $\mathcal{A}$, we may benefit if we guide the generation process to consider the trade-off between fidelity and diversity in the generated images.
The augmented dataset, $\widetilde{\mathcal{X}}$, should contain points \textit{``close but not too close''} to the training data, 
as illustrated in panel (c) of \figureautorefname~\ref{fig:intuition} by the green circles.
To guarantee high-quality images and to avoid artefacts, \ie, high fidelity, the synthetic images should be \textit{close} to the real training images.
To ensure diversity, the generated images should not lie \textit{too close} to the original images, but at some distance from them.
With a DA procedure that increases both fidelity and diversity, our hypothesis is that the downstream model, $\mathcal{M}$, should generalise better.

\subsubsection{Loss function}\label{subsec:LatentAugmentPolicy}

We propose a loss function, $\mathcal{L}(w)$, that takes into account both fidelity and diversity.
The loss function is the weighted sum of four terms, one controlling the fidelity, $\mathcal{L}_f(w)$, and three controlling the diversity, $\mathcal{L}_d (w)$, as
\begin{align}
     \mathcal{L} (w)
        & = \alpha_{f}\mathcal{L}_{f}(w) - \mathcal{L}_d(w) \label{eq:loss-latentaugment}\\
        & = \alpha_{f} \mathcal{L}_{f}(w) \,- \nonumber\\
        &\qquad\quad \underbrace{\big(\alpha_{pix} \mathcal{L}_{pix}(w) + \alpha_{perc} \mathcal{L}_{perc}(w) + \alpha_{lat} \mathcal{L}_{lat}(w)\big)}_{\mathcal{L}_d(w)}, \nonumber
\end{align}
where $\alpha_{f}$, $\alpha_{pix}$, $\alpha_{perc}$, and $\alpha_{lat}$  are positive real weights for the following terms,
\begin{align}
    \mathcal{L}_{f}(w) &= \log\left(1 + \exp^{-D\big(G(w)\big)}\right),\label{eq:disc_loss}  \\
    \mathcal{L}_{pix}(w) &= \frac{1}{N}\sum_{j=1}^N d_{pix}\big(G(w), x_j\big), \label{eq:pix_loss} \\
    \mathcal{L}_{perc}(w) &= \frac{1}{N}\sum_{j=1}^N d_{perc}\big(G(w), x_j\big), \label{eq:perc_loss}\\
    \mathcal{L}_{lat}(w) &= \frac{1}{N}\sum_{j=1}^N d_{lat}(w, w_j^{*}),
    \label{eq:lat_loss}
\end{align}
where $x_j \in \mathcal{X}$ and $w^*_j \in \mathcal{W}$ denote the $j$th training image and its corresponding latent vector (given by the inverse mapping), respectively.
While the next paragraphs detail the four loss terms presented in \equationautorefname{}s~\ref{eq:disc_loss}--\ref{eq:lat_loss}, $\mathcal{L}_d (w)$ consists of three losses measuring different aspects of diversity of $\tilde{x}$ from the real images in $\mathcal{X}$ at different levels of abstraction, including the image space $\mathcal{L}_{pix}(w)$, a perceptual space  $\mathcal{L}_{perc}(w)$, as well as a semantic space $\mathcal{L}_{lat}(w)$.

\subsubsection*{Fidelity Loss}

The fidelity loss, $\mathcal{L}_f(w)$, estimates the fidelity of $\tilde{x}$ as the realness score given by the discriminator, $D$, of the SG2 model (\equationautorefname~\ref{eq:disc_loss}).
A low value suggests that the generated images look unrealistic or that they contain features not present in the real data.
This realness score is based on observing that the discriminator, $D$, distinguishes between real and generated images using low-level and high-level image features automatically learnt for this purpose.
Blau and Michaeli~\cite{blau2018perception} showed that the generator's ability to fool the discriminator, which corresponds to a high realness score, correlates with human opinion scores over synthetic images.

\subsubsection*{Pixel Loss}

The \textit{pixel loss}, $\mathcal{L}_{pix}(w)$, measures how far a generated image, $\tilde{x}$, is from the training images.
Intuitively, large values of $\mathcal{L}_{pix}(w)$ imply larger diversity in the generated images. 
The distance metric, $d_{pix}$, in \equationautorefname~\ref{eq:pix_loss} was the mean squared error between $\tilde{x}$ and a real image, $x_j$, \ie,
\begin{equation}
    d_{pix}\left( G(w), x_j \right)=\frac{1}{ r_{x} c_{x} h_{x} }\lVert G(w) - x_j \rVert_{2}^{2},
\end{equation}
where $r_x$ and $ c_x$ denote the number of rows and columns of the images, respectively, and $h_x$ denotes the number of channels.

\subsubsection*{Perceptual Loss}

The pixel loss does not capture perceptual differences between the generated images and the real images.
For example, consider two images where one of them is a copy of the first but with a spatial offset of one pixel by column. Then, despite their high perceptual similarity, $\mathcal{L}_{pix}(w)$ may be large.
To address this limitation, we measure how far two images are in a feature space by incorporating a perceptual loss, $\mathcal{L}_{perc}(w)$.
To this end, we use the high-level image feature representations extracted from multiple convolutional layers of a VGG network~\cite{simonyan2014very}, that was pretrained on the ImageNet dataset~\cite{russakovsky2015imagenet}. This is the \textit{de facto} standard feature extractor for perceptual losses~\cite{johnson2016perceptual,dosovitskiy2016generating}.
Let $\phi^{l}(x_j)$ denote the activation of the $l$th convolution layer of the VGG network when processing an image, $x_j$.  
The $\phi^{l}(x_j)$ consists of $h_l$ feature (activation) maps of size $r_{l} \times c_{l}$.
We then define the perceptual distance as,
\begin{equation}
    d_{perc}(G(w), x_j)
        = \sum_{l=1}^L \frac{1}{r_{l} c_{l} h_{l}} \big\| \phi^{l}\big(G(w)\big) - \phi^{l}(x_j) \big\|_{2}^{2},
\end{equation}
where $L$ is the total number of layers used. The perceptual distance is averaged over the training images in \equationautorefname~\ref{eq:perc_loss}. 

We computed the perceptual distance $d_{perc}$ considering $64 \times 64$ patches randomly extracted from the real and synthetic images, to save memory, as is common in the literature~\cite{zhang2018unreasonable}.

\subsubsection*{Latent Loss}

The perceptual loss captures the semantic content and overall spatial structure of an image, but uses a feature representation that is general, and may therefore not be optimal for specific applications.
Therefore, to capture specific features in these images, we also incorporated a semantic loss in the latent space of the SG2 model, which has been shown to encode rich semantics~\cite{jahanian2019steerability}.

When two latent vectors, $w_{1}, w_{2} \in \mathcal{W}$, are ``close'' in the latent space, the corresponding images, $x_{1}, x_{2} \in \mathcal{X}$, are semantically similar~\cite{karras2019style}.
The purpose of the \textit{latent loss}, $\mathcal{L}_{lat}(w)$, is to exploit this property and encourage the augmented images to be some distance away from the latent representations of the latent vectors corresponding to images from the training set.
The latent distance, $d_{lat}$, is defined as the mean squared error between $w$ and $w^*_j$, as
\begin{equation}
    {d}_{lat}(w, w_j^{*})= \frac{1}{d_{w}} \lVert w - w_j^{*} \rVert_{2}^{2}.
\end{equation}
where $d_{w}$ is the dimensionality of the latent space.

\subsubsection{Navigating the latent space}

\begin{figure*}[!h]
    \centering
    \includegraphics[width=\textwidth]{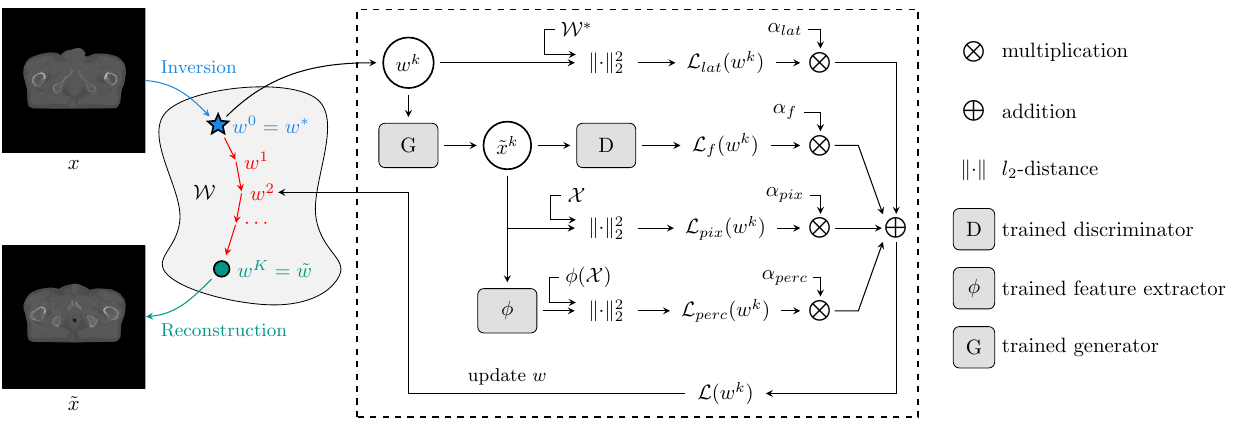}
    \caption{High-level schematic representation  of LatentAugment policy (dashed rectangle). 
    It first retrieves the latent code $w^*$ of the image $x$ we aim to augment.
    Then for each iteration $k$ of the procedure, we manipulate $w^k$ minimising the loss $\mathcal{L} (w^k)$, which is given by a weighted sum of terms computing the fidelity and the  diversity of $x^k$, where the latter is measured at multiple-level (spatial, perceptual and semantic). 
    The minimisation process consists in learning the walk from $w^*$ to the $\tilde{w}$ in an agnostic manner, thus producing the  final augmented image $\Tilde{x}$.
    Note that we used the projection of the real image as a starting point of the policy, \ie, $w^0=w^*$.
    }
    \label{fig:latent_aug}
\end{figure*}

LatentAugment navigates the latent space, $\mathcal{W}$, of the SG2 model by minimising the loss function $\mathcal{L}(w)$,
\begin{equation} \label{eq:search_objective}
    \tilde{w} = \arg \min_{w} \mathcal{L}(w).
\end{equation}
This procedure is illustrated in~\figureautorefname~\ref{fig:latent_aug}, where the dashed rectangle specifies the LatentAugment policy.
To keep a relation between the original and augmented image  ($x$ and $\tilde{x}$),
the starting point of LatentAugment is $w^*$, the latent code retrieved from   $x$. 

In each step, $k$, we have a latent vector, $w^k$,
and use the generator, $G$, to reconstruct a corresponding synthetic image, $\tilde{x}^k$.
The overall loss $\mathcal{L}(w^k)$ is a weighted sum of the four terms, where each weight becomes a hyper-parameter of the policy that determines the relative importance of the different terms, also handling differences in relative scales of the terms.
Different weight magnitudes allow following different directions when navigating the latent space, $\mathcal{W}$, \eg, setting $\alpha_f$ to zero causes the policy to navigate the latent space only with respect to diversity.
The final latent vector, $\tilde{w}$, is input to the generator, $G$,  resulting in a corresponding synthetic image, $\tilde{x}$, that is fed to the downstream model, $\mathcal{M}$, as per \equationautorefname~\ref{eq:AugmentRule}.

\subsection{Hyper-parameter search} \label{subsec:hyper_parameter}

LatentAugment navigates the latent space via a gradient-based optimisation  to find $\Tilde{w}$.
We used the Adam algorithm~\cite{kingma2014adam} with $K$ iterations with learning rate $\eta$. At each step $k$, the current weight vector, $w^k$, is updated as,
\begin{equation}
    w^{k+1} = w^k - \eta \nabla \mathcal{L}(w^k).
\end{equation}
The Adam momentum parameters, $\beta_1$ and $\beta_2$, were set to $0.9$ and $0.999$, respectively.

The hyper-parameters of the proposed approach, namely  $p_{aug}$, $\alpha_{f}$, $\alpha_{pix}$, $\alpha_{perc}$, $\alpha_{lat}$, $K$, and $\eta$, control the GAN generation process.
By fixing $p_{aug}$ and tuning $\alpha_{f}$, $\alpha_{lat}$, $\alpha_{pix}$, and $\alpha_{perc}$, the method specifies the different directions to move on the manifold. 
By tuning the number of optimisation steps, $K$, and the learning rate, $\eta$, the method regulates the intensity of the augmentation on the manifold, and by that how far away the augmented images should be from the real images.
Hence, LatentAugment navigates the latent space without the need for any external supervision that requires human labels or pre-trained models.

We propose two approaches to fine-tune the hyper-parameters, both using the tree-structured Parzen estimator~\cite{bergstra2011algorithms} for $50$ iterations.
The first minimises the Mean Absolute Error (MAE) in the downstream task on the validation set and thus depends on the particular downstream task.
The second maximises the F1~score between real images in the validation set and 50,000  synthetic images generated by LatentAugment.
In the experiments here, the validation set contained $10$k images.
Using the definitions of precision and recall introduced by Kynk{\"a}{\"a}nniemi \textit{et~al.}~\cite{kynkaanniemi2019improved}, which were shown to be well suited to assess both the visual quality and mode coverage of images synthesised by generative models, the F1~score is defined as
\begin{equation}
    \text{F1~score} = 2\cdot\frac{\text{precision} \cdot \text{recall}}{\text{precision} + \text{recall}}.
\end{equation}
In other words, by maximising the F1~score of the SG2, we define a task-agnostic approach that searches for the parameters that make LatentAugment generate both high-quality and diverse synthetic samples.
It is worth noting that such an agnostic approach does not set the value of $p_{aug}$ that, in turn, depends on the specific downstream task. 
Hence, after  maximising the F1~score and setting the  hyper-parameter values, we also   train the downstream model using values of $p_{aug}$ in the range $[0.0, 1.0)$, divided  into $10$ steps.
This allows us to set $p_{aug}$ also for the F1~score fine-tuning, ensuring a fair comparison between the two approaches searching for the hyper-parameters.

\section{Experiments} \label{sec:experiments}

Here we detail the dataset and the downstream task on which the LatentAugment policy was tested.
We then introduce the implementation details for StyleGAN2.
Finally, we describe the experimental comparisons of the DA methods.

\subsection{Dataset and Pre-Processing}

The utility of the described data augmentation methods was evaluated on the downstream application of generating synthetic CT (sCT) images from corresponding Magnetic Resonance (MRI) images. This is an important step in the ambition towards MRI-only radiotherapy~\cite{edmund2017review}.
The data for this example were collected between January 2020 and October 2021 at the University Hospital of Umeå, Umeå, Sweden, from 375 patients (330 male and 45 female) with prostate (243 patients), post-surgery prostate (43 patients), gynaecological (21 patients), rectal/anal (34 male and 22 female patients), and bladder (10 male and 2 female patients) cancer.
The data contained $T_2$-weighted MRI images, captured using a GE Signa 3T PET/MRI scanner (GE Healthcare, Chicago, Illinois, United States; the echo time was approximately 90~ms and the repetition times around 14,000~ms), and corresponding CT scans captured using a Philips Brilliance Big Bore (Philips Medical Systems, Cleveland, OH, USA). 
The images had a resolution of $512\times512$ in $131$ slices per patient, and the slices were subsampled to $256\times256$ using linear interpolation. 
The MRI images were clipped to the range $[0, 1900]$ and the CT images to the range $[-1000, 2000]$ and then normalized to the range $[0, 255]$.
The dataset was split according to a hold-out validation where 70~\% of the images were used for training the SG2 and the $\mathcal{M}$ with each $\mathcal{A}$ procedure, 20~\% were used for validation, and 10~\% were used as a final test set.

\subsection{Downstream Task}

The downstream task that we employed to evaluate the DA procedure was for MRI-to-CT translation, \ie, to generate synthetic CT images from the corresponding MRI images.
For this, we employed the Pix2Pix model~\cite{isola2017image}.
The Pix2Pix model is a straight-forward and computationally efficient image-to-image translation model~\cite{saxena2021comparison} that has demonstrated the ability to generate high-quality images across a variety of tasks~\cite{isola2017image} including MRI-to-CT~\cite{fetty2020investigating}.
The primary objective was to explore whether and to which extent LatentAugment could improve the final model performance when compared with other DA strategies for a realistic and important medical imaging task. The aim was thus not to pinpoint the best MRI-to-CT translation model, but to use the MRI-to-CT task to evaluate the proposed DA procedure.

To ensure a fair comparison, we used the same training strategy and hyper-parameters for Pix2Pix across all the examined DA policies.
We used the default training configuration of the Pix2Pix model (see Isola \textit{et al.}~\cite{isola2017image} for details) except for the number of epochs and the batch size.
We set the number of epochs to 40 and scheduled the learning rate to decay linearly over the last $20$ epochs. The batch size was fixed at $16$.

\subsection{SG2 training} \label{subsec:sg2_training}

In the implemented SG2 model, the generator, $G$, synthesised paired CT--MRI images, each with the resolution $256 \times 256$.
We adhered to the recommended SG2 settings~\cite{karras2020analyzing}, including a batch size of $16$, a mapping network, $\mathcal{F}$, with a depth of two, generator and discriminator learning rates set to $0.0025$, and a regularisation weight of $R_1 = 0.8192$.
The mapping network, $\mathcal{F}$, learned to map $z$, from the $512$-dimensional latent space $\mathcal{Z}$, to the intermediate $512$-dimensional latent space, $\mathcal{W}$.
The SG2 model was trained for 10,000 iterations; processing 1,000 training images in each iteration. 

To avoid overfitting the discriminator, which is common in medical applications with limited data availability, we used the adaptive discriminator augmentation scheme proposed by Karras \textit{et~al.}~\cite{karras2020training}.
It adjusts the probability of applying image-based transformations during SG2 learning, such as translation, shift, \etc\footnote{Refer to Karras \textit{et al.}~\cite{karras2020training} for the list of available transformations.} by a fixed amount according to an overfitting/underfitting heuristic.
The transformations included pixel blitting operations: horizontal flips (xflip) and integer translations (int); and geometric transformations: isotropic scaling (scale), rotation (rotate), anisotropic scaling (aniso), and fractional translation (frac).
The magnitude of each operation was adjusted to avoid generating implausible images, while the probability of performing each transformation was adaptively tuned during the SG2 training~\cite{karras2020training}.
The quality of the generated \textit{paired} CT--MRI images was assessed by comparing to CT and MRI images generated using two SG2 models when trained for each modality separately.
To determine the performance differences due only to the multimodal image generation, we turned off the adaptive discriminator augmentation here.
Then, to understand which image transformation set was best suited for the multimodal training of SG2, we performed a grid search over the transformation space.

We designed a total of six experiments, and repeated each experiment three times, training a total of $18$ SG2 models.
To evaluate the performance, we used the Fréchet Inception Distance (FID)~\cite{heusel2017gans}.
The FID measures the dissimilarity of the densities of two (assumed) Gaussian distributions in the feature space of an Inception-V3 model~\cite{szegedy2016rethinking} (one pre-trained on the ImageNet dataset)~\cite{heusel2017gans}.
The FID score was computed using the whole training data set and 50,000 synthetic images.
In the evaluation, each modality was considered separately, \ie, FID scores were computed for the generated CT and MRI image modalities separately.
Then, we selected the SG2 model with the lower FID score during training.
In the unimodal training, the metric was based on the one modality considered.
Finally, we used the Friedman test~\cite{friedman1937use} to assess whether there were any significant differences between the SG2 configurations in terms of their FID scores.

\subsection{Comparative analysis} \label{subsec:comparative_analysis}

To evaluate the performance of LatentAugment, we compared it to a baseline model, a Pix2Pix model without data augmentation (denoted \textit{Baseline}).
We also compared to the common approach of performing a composition of image transformations (denoted \textit{Standard DA})~\cite{xu2023comprehensive}.
We also compared to generating images directly from randomly sampled intermediate latent vectors (denoted \textit{Standard SG2 DA}), an SG2 augmentation policy common in the literature~\cite{xu2023comprehensive}.
The last two procedures are explained in more detail in the following subsections.

\subsubsection{Standard DA}

Let an image transformation be $T: \mathcal{X} \rightarrow \mathcal{X}$, defined on the input image space $\mathcal{X}$.
Each transformation $T \in \mathbb{T}$ takes a magnitude parameter, $\mu$, that determines the intensity of the operation, \eg, the number of degrees to rotate an image by.
Note that some operations (\eg, horizontal or vertical flips) do not use a magnitude parameter.
Let $\tau$ be a sequence of $N_{\tau}$ image transformations and magnitude parameters, $((T_1, \mu_1), (T_2, \mu_2), \ldots, (T_{N_\tau}, \mu_{N_\tau}))$.
Each operation is applied in sequence, with probability $p_{aug}$.
Hence, the output of $\tau$ is the result of a composition of image transformations, $\tau(x) = T_{N_\tau}(\cdots T_2(T_1(x; \mu_1); \mu_2)\cdots; \mu_{N_\tau})$ and yields an augmented image, $\Tilde{x} = \tau(x)$.
Note that when using this type of augmentation, it is necessary to fine-tune the transformation pipeline.
Indeed, specifying the type, order, and magnitude of the operations is essential to preserve image labels.
 
We used horizontal flips (xflip) and affine/non-affine transformations, which are known to be well-suited in the medical domain~\cite{hussain2017differential}.
Within the affine transformations, we considered rotations (rotate) and fractional translations (frac).
The non-affine transformation considered was elastic deformations (deform)~\cite{simard2003best}.
We set the magnitude range to rotate images (rotate) to $[-3, 3]$ degrees and for translations (frac) to $[-5\%, 5\%]$, meaning the percentage of pixels to shift the image by.
The elastic deformations were implemented with a receptive field of $63$, and the standard deviation of a Gaussian filter was $32$.
The elastic deformations performed a smooth displacement of the pixels in the images exploiting a randomly generated displacement field that was convolved with a Gaussian filter.

\subsubsection{Standard SG2 DA}

Given a latent vector $z$ from $\mathcal{Z}$, sampled from an isometric standard normal prior distribution, the latent vector was mapped to the intermediate latent space, $\mathcal{W}$, by the mapping network, $w=\mathcal{F}(z)$.
Next, the SG2 generator takes the intermediate latent vector, $w$, and generates the corresponding image, $\Tilde{x}=G(w)$.

\subsubsection{Hyper-parameter search} \label{subsec:hyper_parameter_comparison}

The DA approaches were applied according to the augmentation rule in \equationautorefname~\ref{eq:AugmentRule}, where $p_{aug}$ controlled the number of synthetic samples used to train $\mathcal{M}$ in each epoch.
As stated in \sectionautorefname~\ref{sec:methods}, when $p_{aug}=0$ we fed $\mathcal{M}$ only augmented images, while when $p_{aug}=1$ we would only use real images.
For each procedure, we directly augmented the paired CT--MRI images.
For the Standard DA, we carried out an exhaustive search over the image transformation space that consisted of xflip, affine (rotate and frac) and non-affine (deform) transformations.
With a transformation space with $3$ options, there are a total of $7$ combinations, where, for each combination, we allow $p_{aug}$ to assume values from $[0.0, 1.0)$ in ten steps for a total of $70$  experiments.
For the Standard SG2 DA, we conducted a total of $10$ experiments using the Pix2Pix model, for a grid of $p_{aug}$ values in the interval $[0.0, 1.0)$.
For each experiment, we evaluated the downstream model's MAE on the validation set (within the body), searching for the parameter configuration that minimised the MAE.

\subsection{Validation approach}

The metrics used to evaluate the final performance were: the MAE, Structural Similarity (SSIM), Peak Signal-to-Noise Ratio (PSNR), and Learned Perceptual Image Patch Similarity (LPIPS) computed on the test patients.
The MAE was evaluated within the body, excluding air with Hounsfield Unit (HU) values below $-500$.
To assess whether the Pix2Pix models, trained using the three DA policies, performed similarly, we used the Friedman test~\cite{friedman1937use}, and, in case statistically significant differences were detected ($p< 0.05$), we used the Nemenyi posthoc test~\cite{nemenyi1963distribution} to detect pairwise differences between the augmentation policies.
We also evaluated the computational overhead for each DA policy compared to the baseline computing the throughput, \ie, the time (in seconds) needed to augment a batch of $16$ images with a resolution of $256 \times 256$.
In order to have a mean estimate of throughput, we report the mean value registered during the training.
The PyTorch implementation of LatentAugment and all models, are available at \href{https://github.com/ltronchin/LatentAugment}{https://github.com/ltronchin/LatentAugment}.
The experiments were performed using one NVIDIA RTX A5000 GPU.

\section{Results and discussion} \label{sec:results}

In this section, we first discuss the generation of SG2.
Then we present the evaluation of LatentAugment in three experiments: (1) A quantitative and qualitative comparison of the proposed LatentAugment policy to existing DA methods (\subsectionautorefname~\ref{subsec:performance_analysis}), (2) an assessment of LatentAugment sensitivity to hyper-parameter settings (\subsectionautorefname{}s~\ref{subsec:par_importance} and \ref{subsec:ablation}), and (3) an exploration of LatentAugment ability to tackle the generative learning trilemma (\subsectionautorefname~\ref{subsec:trilemma}).

\subsection{SG2 image generation assessment}

\begin{figure}[!h]
     \centering
     \begin{subfigure}[b]{0.49\columnwidth}
         \centering
         \includegraphics[width=\columnwidth]{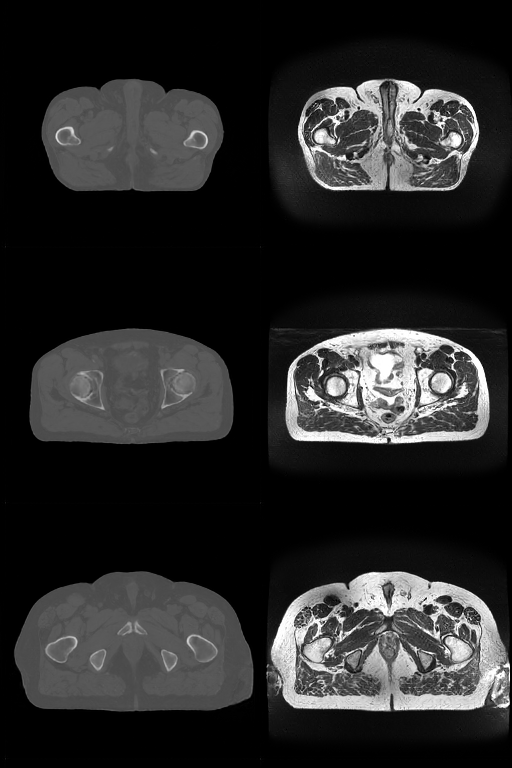}
         \caption{Synthetic sample (CT--MRI).}
         \label{subfig:synthetic}
     \end{subfigure}
     %\hfill
     \begin{subfigure}[b]{0.49\columnwidth}
         \centering
         \includegraphics[width=\columnwidth]{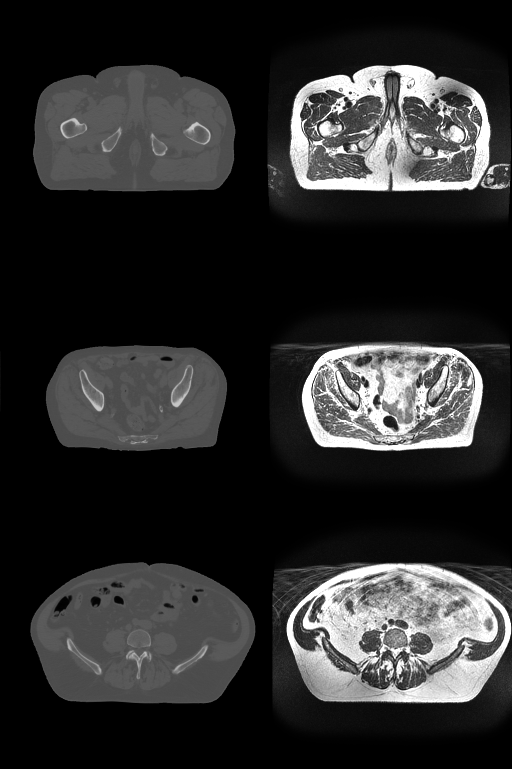}
         \caption{Real samples (CT--MRI).}
         \label{subfig:real}
     \end{subfigure}
     \caption{Left: Random synthetic samples generated by SG2. Right: Real images are randomly drawn from the training set. 
     }
     \label{fig:visual_inspection}
\end{figure}

In \tableautorefname~\ref{tab:SG2-training} we show the results from the different SG2 settings detailed in \subsectionautorefname~\ref{subsec:sg2_training}.
The table is organised in two sections.
In the first, the SG2 was trained without DA: the first and second rows show the FID scores in the case of unimodal training, whilst the third row corresponds to multimodal training.
These results  show that multimodal training leads to reduced performance compared to unimodal models: this could be expected since the multimodal SG2 has to learn a more complex data manifold comprising two modalities.
The second section of~\tableautorefname~\ref{tab:SG2-training} shows FID scores when the SG2 was trained with three sets of geometric transformations.
When comparing these three lines  against the third row of the previous section, we notice that DA allows decreasing FID values, which correspond to better performance.
In particular, when augmenting the data by scaling, rotation, anisotropic scaling,
and fractional translation  (fifth row in \tableautorefname~\ref{tab:SG2-training}) we get  the lowest FID scores for the multimodal training, with a mean of $8.11$ (between the FIDs for CT and MRI).
Hence, in all the next  experiments using DA  we employ the SG2 generator trained with such four transformations; an example of its generation is shown in \figureautorefname~\ref{fig:visual_inspection}.
Moreover, the Friedman test reveals that no significant differences between any of the experiments reported in \tableautorefname~\ref{tab:SG2-training}) exist ($p = 0.45$).

\begin{table}[]
\centering
\resizebox{\columnwidth}{!}{%
\begin{tabular}{@{}ccccclcc@{}}
\toprule
\multicolumn{6}{c}{Adaptive discriminator augmentation} & \multicolumn{2}{c}{FID $\downarrow$} \\ \midrule
xflip & int & scale & rotate & aniso & frac & CT & MRI \\ \midrule
\xmark & \xmark & \xmark & \xmark & \xmark & \xmark & \bm{$6.79 \pm 0.47$} &  \\
\xmark & \xmark & \xmark & \xmark & \xmark & \xmark &  & $8.81 \pm 0.32$ \\
\xmark & \xmark & \xmark & \xmark & \xmark & \xmark & $14.97 \pm 0.78$ & $10.44 \pm 0.68$ \\ \midrule
\cmark & \cmark & \xmark & \xmark & \xmark & \xmark & $11.58 \pm 0.83$ & $8.04 \pm 0.52$ \\
 \xmark& \xmark & \cmark & \cmark & \cmark & \cmark & $9.10 \pm 0.71$ & \bm{$7.12 \pm 0.24$} \\
\cmark & \cmark & \cmark & \cmark & \cmark & \cmark & $10.62 \pm 1.07$ & $11.35 \pm 0.64$ \\ \bottomrule
\end{tabular}%
}
\caption{The first three rows report the average and the standard error of FID values achieved during the unimodal and multimodal setting of SG2. 
    Lower FID scores indicate higher-quality images.
    The last three rows show FID scores for three sets of geometric transformations used during SG2 training (acronyms are defined in subsection~\ref{subsec:sg2_training}).
    The best results are reported in bold.}
\label{tab:SG2-training}
\end{table}

\subsection{Analysis of the downstream task performance} \label{subsec:performance_analysis}

\begin{table*}[]
    \centering
    %\resizebox{\textwidth}{!}{%
    \begin{tabulary}{\textwidth}{@{}lcccccc@{}}
    \toprule
    Method & \multicolumn{1}{l}{$\text{Objective}^{\text{*}}$} & MAE $\downarrow$ & SSIM $\uparrow$ & PSNR $\uparrow$ & LPIPS $\downarrow$ & $\text{Throughput}^{\text{**}} [\text{sec}]\downarrow$ \\ \midrule
    Baseline &  & $45.64$ &  $9.29 \cdot 10^{-1}$ & $33.62$ & 
    $6.57 \cdot 10^{-2}$ & \\
    Standard DA & MAE & $44.73$ &  $9.31 \cdot 10^{-1}$ & $33.82$ & $6.32 \cdot 10^{-2}$ & $3.48 \cdot 10^{-2}$ \\
    Standard SG2 DA & MAE & $42.26$ &  $9.34 \cdot 10^{-1}$ & $34.19$ & $6.45 \cdot 10^{-2}$ & \bm{$3.09 \cdot 10^{-2}$} \\ \midrule
    \multirow{2}{*}{LatentAugment} & F1~score & \bm{$39.32$} & \bm{ $9.37 \cdot 10^{-1}$} & $34.29$ & \bm{$6.10 \cdot 10^{-2}$} & \multirow{2}{*}{$2.55$} \\
     & MAE & $39.39$ &  $9.37 \cdot 10^{-1}$ & \bm{$34.41$} & $6.33 \cdot 10^{-2}$ &  \\ \bottomrule
    \end{tabulary}%
    %}
    \caption{Downstream task performance on the test set. 
    We marked each metric with a lower or upper arrow to define whether it has to be minimised or maximised.
    The best results for each metric are reported in bold. 
    *: Objective defines the metric used to perform the hyper-optimisation procedure of each DA policy on the validation set.
    **: Throughput is computed on a
    single NVIDIA RTX A5000, with a batch size equal to 16 and
    an image resolution equal to $256 \times 256 \times 2$.
    }
    \label{tab:performances}
\end{table*}

The LatentAugment hyper-parameter search presented in \subsectionautorefname~\ref{subsec:hyper_parameter} returned two sets of values. 
The first minimised the MAE on the validation set, and it consists of $p_{aug}=0.7$, $\alpha_{pix}=3$,  $\alpha_{perc}=1$, $\alpha_{lat}=0.1$.
The second, sought to maximise the F1~score between synthetic and the real images, and it set  $\alpha_{pix}=0.1$, $\alpha_{perc}=10$, $\alpha_{lat}=0.001$, $\alpha_{f}=0.01$, $K=9$, $\eta=0.01$.
Let us recall that the hyper-parameter search based on F1~score does not include $p_{aug}$ because it does not depend on a downstream task, as described in subsection~\ref{subsec:hyper_parameter}: 
the additional grid search on $p_{aug}$ for the F1~score fine~tuning sets $p_{aug} = 0.8$. 
It is worth noting that despite targeting different metrics, both methods set the same values for the hyper-parameters controlling the intensity of the augmentation on the manifold ($K$ and $\eta$).
This underscores a potential correlation between the inherent diversity of synthetic images and their efficacy in DA applications: the more diverse the synthetic images, the more effective the DA.

Let us now turn our attention to hyper-parameter searches for Standard DA and Standard SG2 DA, which are the other approaches used for comparison (\subsectionautorefname~\ref{subsec:comparative_analysis}).
For Standard DA we find that the best set of hyper-parameters includes  $p_{aug}=0.9$, xflip, affine (rotate and frac), and non-affine (deform).
For  Standard SG2 DA, $p_{aug}=0.7$ performed the best.

It is interesting to notice that $p_{aug}$, controlling the probability of using augmented images,  falls between $0.7$ and $1.0$ in all the four hyper-parameter searches, two for LatentAugment,   one for Standard DA, and another for Standard SG2 DA  cases.
This suggests that training the downstream model with at least $70\%$ real images ensures increased performance.

\tableautorefname~\ref{tab:performances} presents downstream task results by summarising the MRI-to-CT translation performance on the test set for all the DA methods.
For each DA, it shows, the objective optimised by the hyper-parameter search on the validation set, the performance metrics (MAE, SSIM, PSNR, and LPIPS), the time required to augment a batch of $16$ samples with resolution $256 \times 256$ (the throughput).
It is worth noting that the proposed LatentAugment method performs better than the Baseline by a large margin with both objectives.
With the F1~score objective, LatentAugment achieves a $13.8\%$ decrease in MAE, $0.9\%$ and $2.0\%$ increases in SSIM and PSNR respectively, and a $7.1\%$ decrease in LPIPS relative to Baseline.
The MAE objective resulted in similar performance gains, except for a smaller decrease in LPIPS ($3.7\%$ decrease compared to Baseline).
The difference in LPIPS between the two LatentAugment settings is likely due to the search procedure that rewards pixel-based differences (MAE objective), compared to perceptual differences (F1~score objective).
This can also be explained by looking at the values of $\alpha_{perc}$.
Indeed, with the MAE objective $\alpha_{perc}$ is an order of magnitude smaller than the one retrieved by the F1~score objective, \ie, 1 vs 10.
When comparing the results of Standard DA and Standard SG2 DA (lines 2 and 3 in \tableautorefname~\ref{tab:performances}) against LatentAugment, we argue that the latter is able to generate synthetic images that allow  $\mathcal{M}$ to generalise better.
We also notice that both Standard DA and Standard SG2 DA perform better than the Baseline.
However, the performance gains are smaller than the ones achieved by LatentAugment.
Indeed, on the one hand, Standard DA can produce only a limited set of possible sources of variation, relying on the image primitives included in the transformations set, while, on the other hand, Standard SG2 DA lacks control over the generation process, suffering from poor mode coverage.

The last column of \tableautorefname~\ref{tab:performances} reveals that the sampling process of LatentAugment has a larger throughput when creating a batch of augmented images than the Standard DA and the Standard SG2 DA, as it will be discussed in section~\ref{subsec:trilemma}.

\begin{table}[]
    \centering
    \begin{tabulary}{\textwidth}{llccccc p{0em} r}
     & & \rotatebox{90}{Baseline} & \rotatebox{90}{Standard DA} & \rotatebox{90}{Standard SG2 DA} & \rotatebox{90}{$\text{LatentAugment}^{\text{*}}$} & \rotatebox{90}{$\text{LatentAugment}^{\text{**}}$} && \rotatebox{90}{Score} \\ \toprule
    \multirow{5}{*}{\rotatebox{90}{MAE}} & Baseline &  & $0$ & $-$ & $-$ & $-$ && $-3$\\
     & Standard DA & $0$ &  & $-$ & $-$ & $-$ && $-3$ \\
     & Standard SG2 DA & $+$ & $+$ &  & $-$ & $-$ && $0$ \\
     & $\text{LatentAugment}^{\text{*}}$ & $+$ & $+$ & $+$ & & $0$ && $+3$\\
     & $\text{LatentAugment}^{\text{**}}$ & $+$ & $+$ & $+$ & $0$ & && $+3$ \\ \cmidrule{1-7} \cmidrule{9-9}
    \multirow{5}{*}{\rotatebox{90}{SSIM}} & Baseline &  & $0$ & $-$ & $-$ & $-$ && $-3$ \\
     & Standard DA & $0$ &  & $-$ & $-$ & $-$ && $-3$ \\
     & Standard SG2 DA & $+$ & $+$ &  & $-$ & $-$ && $0$\\
     & $\text{LatentAugment}^{\text{*}}$ & $+$ & $+$ & $+$ &  & $0$ && $+3$ \\
     & $\text{LatentAugment}^{\text{**}}$ & $+$ & $+$ & $+$ & $0$ &  && $+3$ \\ \cmidrule{1-7} \cmidrule{9-9}
     \multirow{5}{*}{\rotatebox{90}{PSNR}} & Baseline &  & $0$ & $-$ & $-$ & $-$ && $-3$ \\
     & Standard DA & $0$ &  & $-$ & $-$ & $-$ && $-3$ \\
     & Standard SG2 DA & $+$ & $+$ &  & $-$ & $0$ && $+1$\\
     & $\text{LatentAugment}^{\text{*}}$ & $+$ & $+$ & $+$ &  & $0$ && $+3$ \\
     & $\text{LatentAugment}^{\text{**}}$ & $+$ & $+$ & $0$ & $0$ & && $+2$  \\ \cmidrule{1-7} \cmidrule{9-9}
    \multirow{5}{*}{\rotatebox{90}{LPIPS}} & Baseline &  & $-$ & $0$ & $-$ & $-$ && $-3$ \\
     & Standard DA & $+$ &  & $0$ & $0$ & $-$ && $0$\\
     & Standard SG2 DA & $0$ & $0$ &  & $0$ & $-$ && $-1$\\
     & $\text{LatentAugment}^{\text{*}}$ & $+$ & $0$ & $0$ &  & $-$ && $+1$\\
     & $\text{LatentAugment}^{\text{**}}$ & $+$ & $+$ & $+$ & $+$ &  && $+4$  \\ \bottomrule
    \end{tabulary}%
    \caption{Results of the Nemenyi post-hoc test for MAE, SSIM, PSRN, LPIPS values in \tableautorefname~\ref{tab:performances} when comparing all the DA methods.
    %The symbols minus ($-$), zero ($0$), and plus ($+$) represent significant lower ranking, no significant difference, and significant higher ranking, when comparing a method in the rows to a method in the columns, respectively.
    A zero (0) means non-significant difference, and a minus $-$ (plus $+$) means ranked  statistically significantly lower (higher) when comparing a method in a row to a method in a column.
    *: LatentAugment using a set of parameters optimised on MAE. **: LatentAugment using a set of parameters optimised on F1~score.}
    \label{tab:statistical}
\end{table}

The Friedman test performed for each metric detects significant differences among the methods for all metrics (p-value $<$ 0.05).
We therefore applied the Nemenyi post-hoc test, whose results comparing all pairs of DA methods are shown in \tableautorefname~\ref{tab:statistical}.
When comparing pairs of methods, the symbol minus ($-$)  indicates that a method in a row has a performance ranking that is statistically significantly lower than a method in a column. 
Straightforwardly, the symbols zero ($0$), and plus ($+$)  correspond to no significant difference and statistically significantly higher ranking, respectively.
The table also includes a last column reporting a score computed as the sum of Nemenyi directions,  where $+$ and $-$ mean adding and subtracting one unit, respectively: hence, the larger the values, the more times each DA method wins over the others.
The score ranges from $-4$ to $+4$: a negative value of $-4$ indicates that the DA approach in the row always loses against all the methods in the columns, whilst
 $+4$ denotes the opposite situation. 
The first two lines of each metric in \tableautorefname~\ref{tab:statistical} reveal no significant difference between Standard DA and the Baseline.
This suggests that applying standard image transformations like rotation and shifts does not significantly improve the performance on the MRI-to-CT translation task.
Moreover, when comparing the third line against the first and last two lines for each metric, we find Standard SG2 DA to be significantly better than both the Baseline and the Standard DA, but significantly worse than LatentAugment.
Notably, LatentAugment outperforms the other DA methods.
There is no significant difference in the LatentAugment scores for the two objectives (last two lines for each metric), except for the LPIPS metric, where the F1~score objective performs significantly better than the MAE objective.

\begin{table*}[]
    \centering
    \resizebox{\textwidth}{!}{%
    \begin{tabular}{@{}c|c|cl|cl|clcc@{}}
    \cmidrule(r){1-9}
    \diagbox{CT, MRI}{sCT, \bm{$\Delta$}}{Method}  & Baseline & \multicolumn{2}{c|}{Standard DA} & \multicolumn{2}{c|}{Standard SG2 DA} & \multicolumn{3}{c}{$\text{LatentAugment (ours)}^{\text{*}}$} & \multirow{3}{*}{\includegraphics[scale=0.5]{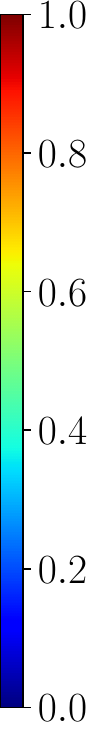}} \\ \cmidrule(r){1-9}
    \ \includegraphics[scale=0.3]{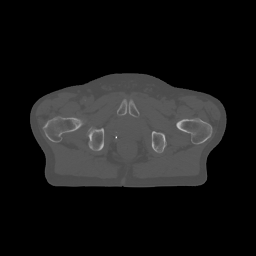} & \includegraphics[scale=0.3]{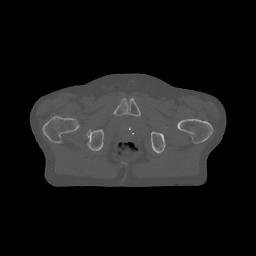} & \multicolumn{2}{c|}{\includegraphics[scale=0.3]{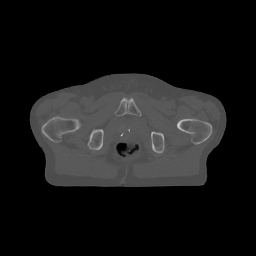}} & \multicolumn{2}{c|}{\includegraphics[scale=0.3]{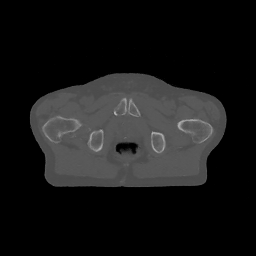}} & \multicolumn{2}{c}{\includegraphics[scale=0.3]{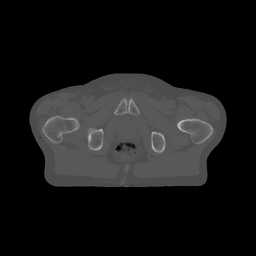}} & \includegraphics[scale=0.3]{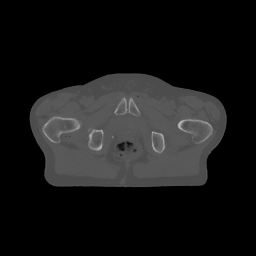} & \\
    ~\includegraphics[scale=0.3]{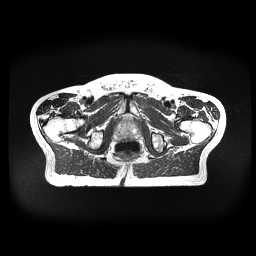} & \includegraphics[scale=0.3]{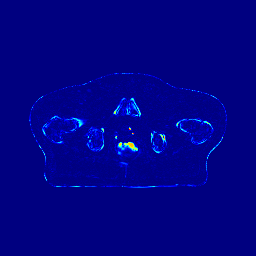} & \multicolumn{2}{c|}{\includegraphics[scale=0.3]{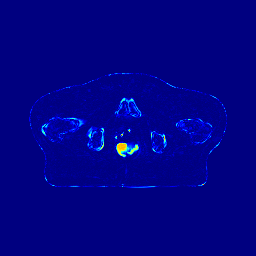}} & \multicolumn{2}{c|}{\includegraphics[scale=0.3]{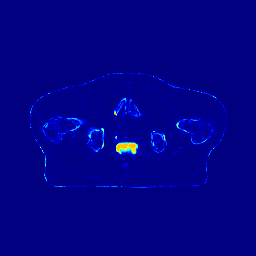}} & \multicolumn{2}{c}{\includegraphics[scale=0.3]{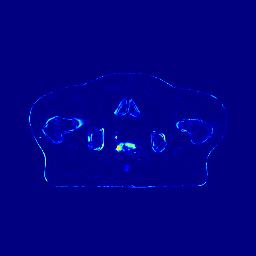}} & \includegraphics[scale=0.3]{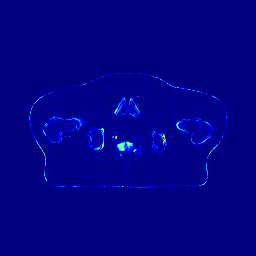} &  \\ \cmidrule(r){1-9}
    \end{tabular}%
    }
    \caption{Qualitative comparisons between DA methods.
    The first column reports the reference CT and MRI test images, respectively.
    Each column denotes a different DA used to train $\mathcal{M}$.
    For each DA method, we report the sCT along with the heatmaps $\Delta=\text{CT}-\text{sCT}$.
    *: left column reports LatentAugment results using a set of parameters optimised on MAE. 
    The right column reports LatentAugment results  using a set of hyper-parameters optimised on F1~score.}
    \label{tab:visual_inspection}
\end{table*}

\begin{figure}[!t]
    \centerline{\includegraphics[width=\columnwidth]{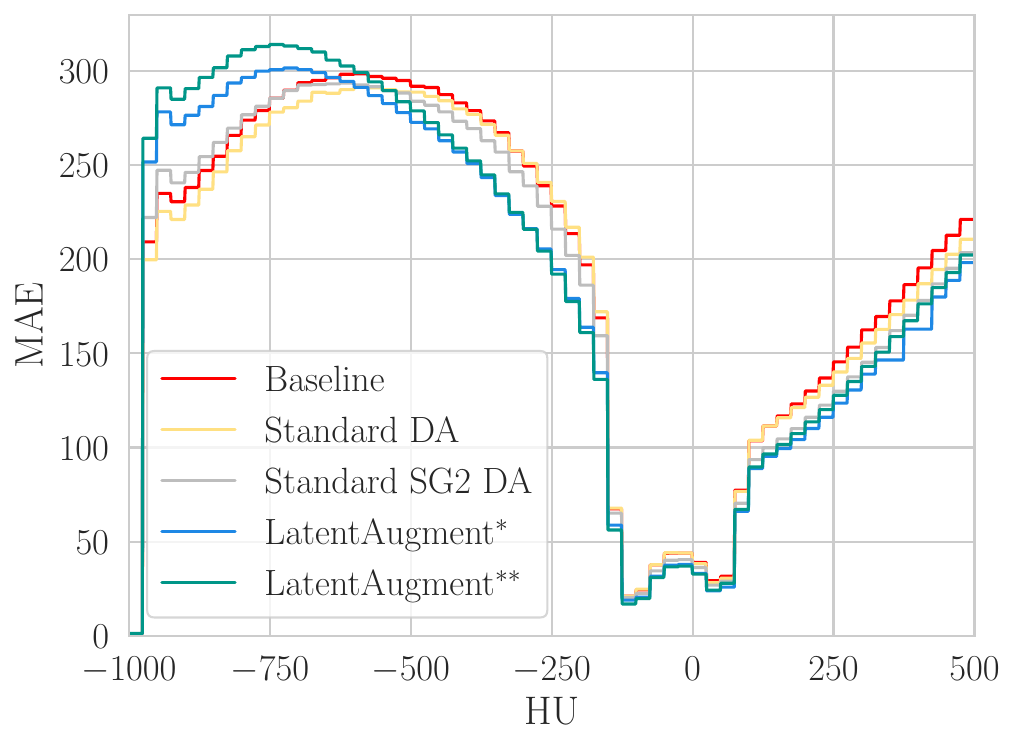}}
    \caption{An illustration of the downstream model performance of the MAE as a function of HU.
    Values between $-350$ HU to $350$ HU denotes soft tissues. Values below $-500$ and those above about $500$ are air and bones, respectively.
    *: LatentAugment with the MAE objective.
    **: LatentAugment with the F1~score objective.
    }
    \label{fig:hu_analysis}
\end{figure}

The qualitative assessment of the different DA methods is illustrated in \tableautorefname~\ref{tab:visual_inspection} and \figureautorefname~\ref{fig:hu_analysis}.
\tableautorefname~\ref{tab:visual_inspection} presents the sCT images generated from an example MRI image when using each DA method.
The reference CT and MRI images are in the first column, and the differences between the ground truth CT and the generated sCT images ($\Delta=\text{CT} - \text{sCT}$) are in the following columns. The heatmaps were normalised to a range from 0 (no error) to 1 (maximum error), and illustrate the errors in the reconstructed sCT images for each method.
From this visual analysis, the Pix2Pix trained without DA (Baseline) demonstrates a higher bone reconstruction error, while Standard DA and Standard SG2 DA yield more accurate reconstruction in the same regions.
All tested DA methods, except LatentAugment, fail to accurately reconstruct the urinary bladder from the MRI image.

In \figureautorefname~\ref{fig:hu_analysis}, we illustrate the mean MAE test set error at different HU values.
HU is a quantitative scale that describes the radiodensity in medical CT and provides an accurate tissue type density.
Air is represented by a value of $-1000$ and bone between $500$ (cancellous bone) to $2000$ (dense bone).
Values centred around $0$ and within an interval of $350$ denote soft tissues.
We observe that LatentAugment has a smaller MAE for both soft tissue and bone while having a worse performance for HU values smaller than $-500$: however, such values indicate areas outside the body and, therefore, are not of interest in the task of MRI-to-CT translation.
Thus, these values are not included  in the computation of the MAE when it is used as a metric. 

\subsection{On the importance of the hyper-parameters} \label{subsec:par_importance}

\begin{figure}[!t]
    \centerline{\includegraphics[width=\columnwidth]{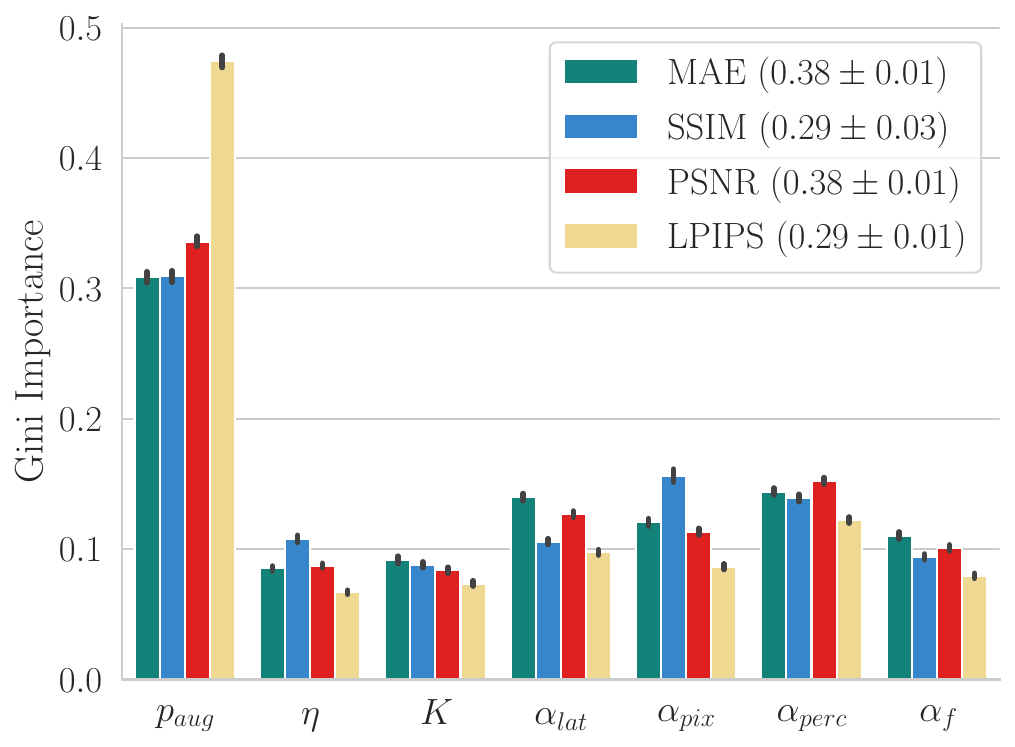}}
    \caption{
    Impact of each LatentAugment hyper-parameter (on the x-axis)  in terms of Gini importance: each bar is a metric
    Each bar for each metric is the mean hyper-parameter Gini importance, whilst the error bars are bootstrapped standard errors.
    Values in round parentheses in the legend are the mean NRMSE, and the related bootstrapped standard error.}
    \label{fig:importance_plot}
\end{figure}

Here we aim to determine the sensitivity of the downstream model performance to LatentAugment hyper-parameters.
This analysis not only clarifies the parameters that influence the success of the augmentation, \ie, better downstream model performance but also provides a hint on how to set them.
To this goal, we trained a random forest regressor to predict the value of a performance metric for each of the $50$ combinations of the hyper-parameter search having MAE as the objective. 
Indeed, we excluded the experiments minimising the F1~score as they do not directly consider the downstream model performance.
Each combination is a sample for the regressor, and it is represented in $\mathbb{R}^7$ because there are $7$ LatentAugment hyper-parameters (\ie, $p_{aug}$, $\eta$, $K$, $\alpha_{lat}$, $\alpha_{pix}$, $\alpha_{perc}$, and $\alpha_{f}$). 
The ground truth is the MAE  obtained by evaluating the downstream model on the validation set. 
This procedure was repeated for the other three metrics, \ie  SSIM, PSNR, and LPIPS.
We used a random forest as a regression algorithm for its capability to handle complex data, enhance prediction accuracy, and provide interpretability through feature importance~\cite{breiman2001random}.
We trained the random forest using $1,000$ bootstrap rounds and, then, it provides a measure of importance for each feature. 
This score is computed as the mean impurity reduction that it brought, and it is usually named Gini importance~\cite{nembrini2018revival}.
We also computed the Normalised Root Mean Squared Errors (NRMSE) on the test set.
Such values are reported in round parenthesis in the legend of~\figureautorefname~\ref{fig:importance_plot}.
The same figure shows the results of the features' importance study: the bars are the mean of the Gini importance of each hyper-parameter for each metric, and the error bars correspond to bootstrapped standard errors. 
The taller the bar, the more important the parameter is for predicting the metric.
We observe that $p_{aug}$ is the most important hyper-parameter: we speculate that this happens since it primarily controls each DA policy and needs to be carefully tuned.

\begin{figure}[!t]
    \centerline{\includegraphics[width=\columnwidth]{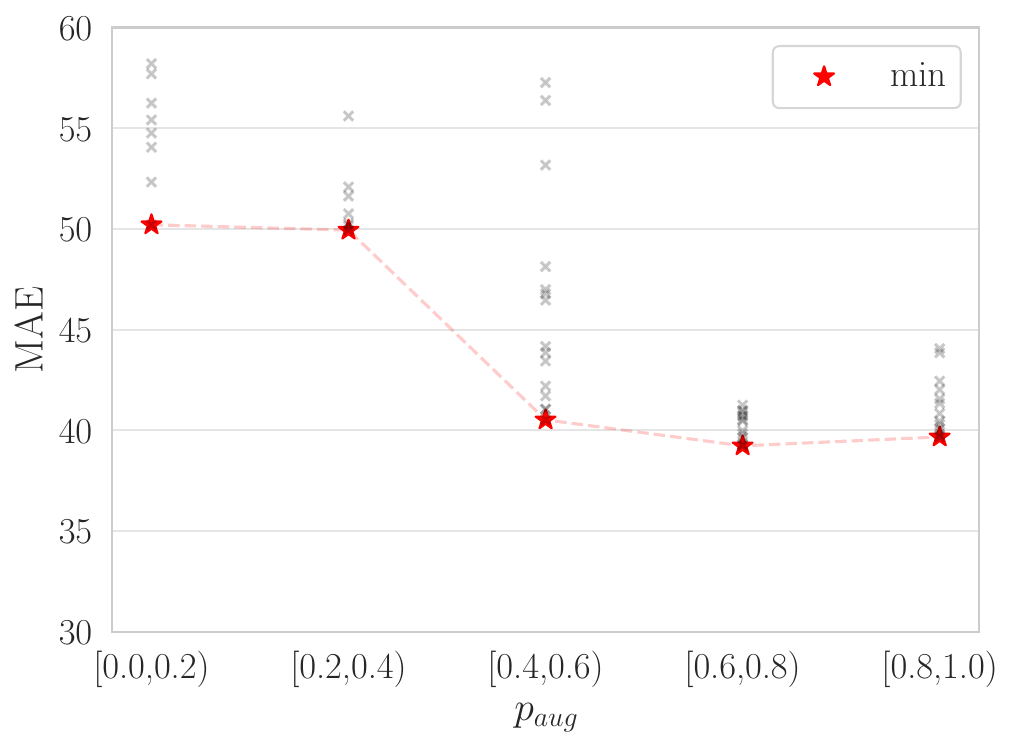}}
    \caption{Effect of $p_{aug}$ on downstream model $\mathcal{M}$. We denote as red start the minimum MAE value computed in each interval and with the grey $\times$ all the experiments that have a $p_{aug}$ in the considered interval.
    We divide the possible range of $p_{aug}$ values in intervals of $0.2$.}
    \label{fig:p_aug_effect}
\end{figure}

To deepen the analysis on $p_{aug}$ and answer the question "How does the amount of augmented data added affect the DA's improvement?" we show in 
\figureautorefname~\ref{fig:p_aug_effect} the validation MAE score obtained across the $50$ hyper-parameter search experiments as a function of $p_{aug}$.
The minimum MAE score in each $p_{aug}$ range is represented by a red star and is the one that a hyper-parameter search procedure would have selected.
A grey $\ times$ marks the MAE values from all the other experiments.
The minimum MAE score was obtained in the interval $[0.6,0.8)$, indicating that some 60\% to 80\% of the training images should be real for a good result.
Indeed, lower values of $p_{aug}$ (stronger augmentation) involve higher amounts of augmented data in the training of $\mathcal{M}$ with the risk of harnessing the stability of the training, \ie, if the Pix2Pix discriminator never sees what the training images really look like, it is not clear if it can guide the generator properly in generating non leaked synthetic images. 
In practical terms, the downstream model might learn the noise in the augmented images as part of the real data distribution, which could result in the noise from the MRI being mistakenly translated to the sCT as well.
On the other hand, using too many real images, \ie, higher values of  $p_{aug}$ may result in not making $\mathcal{M}$ in seeing enough synthetic samples to generalise better.
This also reinforces the observation that the dataset should contain at least $70\%$ real images when training $\mathcal{M}$ for all the DA methods  (\subsectionautorefname~\ref{subsec:performance_analysis}).

Turning out the attention to the importance of the regularisation weights, \figureautorefname~\ref{fig:importance_plot} seems not to highlight any difference in the importance of each parameter.
The diversity terms $\alpha_{lat}$, $\alpha_{pix}$, and $\alpha_{perc}$ appear to be more important than the fidelity term, $\alpha_f$, highlighting that the diversity plays a crucial role in making the augmentation effective for the downstream task. 
Moreover, the regularisation weights appear to be more important than the intensity of the transformations, adjusted by $\eta$ and $K$.
Thus, we argue that it is more important to define useful augmentation direction in the latent space, to the intensity of the transformations.

\subsection{Ablation study} \label{subsec:ablation}

\begin{table}[]
    \centering
    \begin{tabulary}{\columnwidth}{@{}ccc c cc@{}}
    \toprule
    $\alpha_{pix}$ & $\alpha_{perc}$ & $\alpha_{lat}$ & $\alpha_{f}$ & MAE $\downarrow$ & $\text{Throughput}^{\text{*}} [\text{sec}]$ $\downarrow$ \\ \midrule
    $3$ & $1$ & $0.1$ & $0.01$ & $39.39$ & $2.55$ \\ \midrule
    {0} & {0} & {0} & $0.01$ & $ 41.99~(+6.6~\%)$ & $1.91~(-25.1~\%)$ \\
    $3$ & $1$ & {0} & $0.01$ & $41.29~(+4.8~\%)$ & $2.54~(-0.3~\%)$ \\ 
    $3$ & {0} & $0.1$ & $0.01$ & $40.46~(+2.7~\%)$ & $1.97~(-22.9~\%)$ \\
    {0} & $1$ & $0.1$ & $0.01$ & $39.92~(+1.3~\%)$ & $2.50~(-1.9~\%)~$ \\ \midrule
    $3$ & $1$ & $0.1$ & {0} & $40.05~(+1.7~\%)$ & $2.07~(-18.8~\%)$ \\ \bottomrule
    \end{tabulary}%
    \caption{
    Results from the ablation test.
    The $p_{aug}$, $\eta$, and $K$ are set to $0.7$, $0.01$ and $9$, respectively, according to the best configuration found on the validation set minimising MAE.
    The first row denotes the MAE obtained on the test set training $\mathcal{M}$ with the non-perturbed parameter configuration for LatentAugment.
    Rows two to five report the results in disrupting each diversity loss term:  if a weight is set to $0$, the corresponding loss term in the \equationautorefname~\ref{eq:loss-latentaugment} is not used in the augmentation policy.
    The last row removes the fidelity term, $\alpha_f$, while keeping the diversity terms.
    *: Throughput is computed as in \tablename~\ref{tab:performances}. %on a single NVIDIA RTX A5000, with a batch size equal to $16$ and an image resolution of $256 \times 256$.
    }
    \label{tab:ablation_exp}
\end{table}

We now discuss deactivating the diversity-fidelity terms in LatentAugment.
To this end, we start from the best parameter configuration found running the MAE-based hyper-parameter search on the validation set, \ie, $K=9$, $\eta=0.01$, $\alpha_{lat}=0.1$, $\alpha_{pix}=3$, $\alpha_{perc}=1$, and $\alpha_{f}=0.01$ as it directly correlates with the downstream model performance. 
From this starting point, we performed five additional experiments training $\mathcal{M}$ using a perturbed parameter set for LatentAugment.
In the first four experiments, we focused on the diversity weights $\alpha_{pix}, \alpha_{perc}, \alpha_{lat}$, while
in the last experiment, we turned off the fidelity weight, $\alpha_{f}$. 

The results are in the \tableautorefname~\ref{tab:ablation_exp}: the first four columns report the diversity and fidelity weights, while the last two show the MAE achieved on the test set and the throughput, respectively.
Round parenthesis in each tabular denotes the percentage of MAE and throughput variation when ablating the diversity-fidelity terms compared to the references hyper-parameters set (first line in the table).
We achieved the most significant increase in MAE when we deactivated all the diversity loss terms by setting $\alpha_{lat}$, $\alpha_{pix}$, and $\alpha_{perc}$ to $0$ ($6.6\%$ of MAE increase).
Without any regularisation that ensures the augmented images are diverse, we only seek the latent space for the direction that ensures fidelity, a desideratum already satisfied by the SG2's generated images without performing any editing policy. 
In other words, if we do not search the latent space for diversity, we do not need to set a condition to maintain the synthetic images in the manifold of the real data.
Indeed, the difference is very small between the MAE obtained within this configuration ($41.99$ in the second line of \tableautorefname~\ref{tab:ablation_exp}) and the Standard SG2 DA ($42.26$ in the third line of \tableautorefname~\ref{tab:performances}).
The sampling time decreases substantially (by $25.1\%$) without the diversity terms.
To determine the diversity terms that cause the MAE drop, we performed three additional experiments by turning off one diversity term at a time. 
By observing lines three, four, and five of the table, we notice that latent loss is the most important term.
Indeed removing the latent loss ($\alpha_{lat}=0$), produces an MAE increase of $4.8\%$ compared to an MAE increase of  $2.7\%$ and $1.3\%$ when ablating the perceptual loss ($\alpha_{perc}=0$) and the pixel loss ($\alpha_{pix}=0$), respectively. 
This result supports our hypothesis that the highly-structured semantic hierarchy in deep generative representations can be exploited to develop a DA method that manipulates generative models' latent space. 
Moreover, removing the latent loss does not cause a substantial reduction in the throughput, confirming the suitability of working in the latent space from a computational point of view.
In contrast, perceptual loss is the most time-expensive term requiring a forward pass through the VGG network.

Ablating the fidelity loss ($\alpha_{f}=0$) increased the MAE by $1.7\%$: this suggests that providing high-quality images for DA purposes is less important and confirms the analysis carried out in \subsectionautorefname~\ref{subsec:par_importance}.
However, we hypothesise that such a small variation in performance could also be related to corresponding low values of the intensity terms, $\eta$ and $K$, which reduces the importance of the regularisation weights.

\subsection{Tackling the generative learning trilemma} \label{subsec:trilemma}

\begin{figure*}[!h]
     \centering
     \begin{subfigure}[b]{0.49\textwidth}
         \centering
         \includegraphics[width=\textwidth]{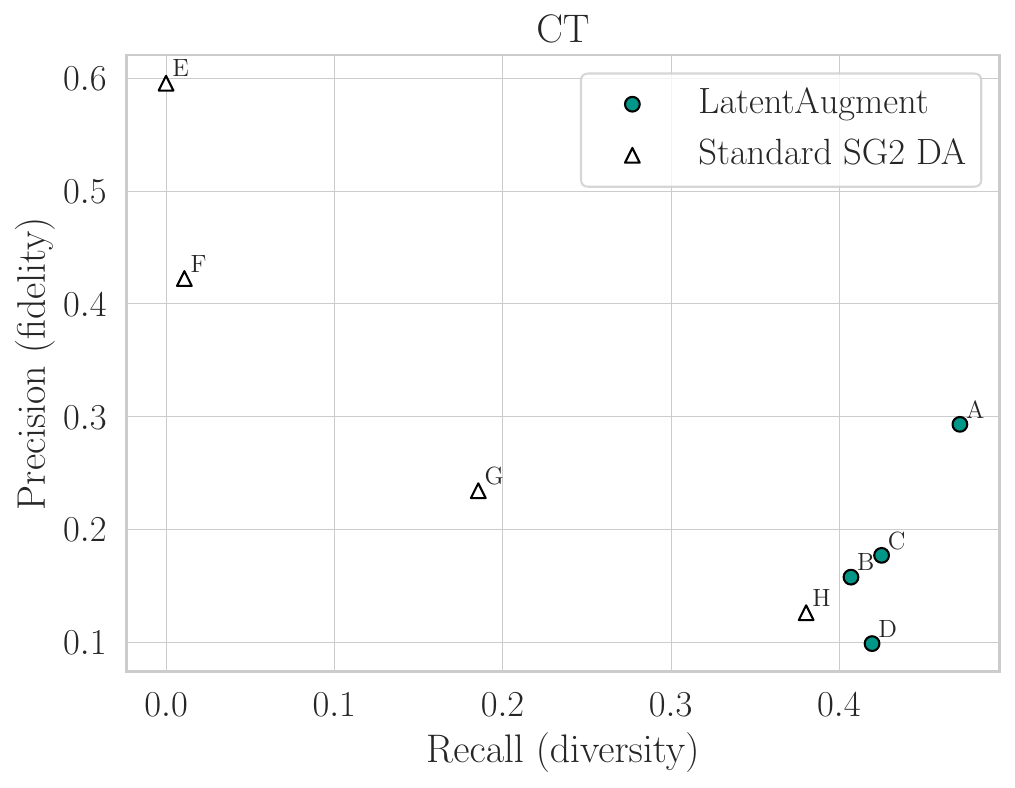}
         \label{fig:pr_ct}
     \end{subfigure}
    \centering
     \begin{subfigure}[b]{0.49\textwidth}
         \centering
         \includegraphics[width=\textwidth]{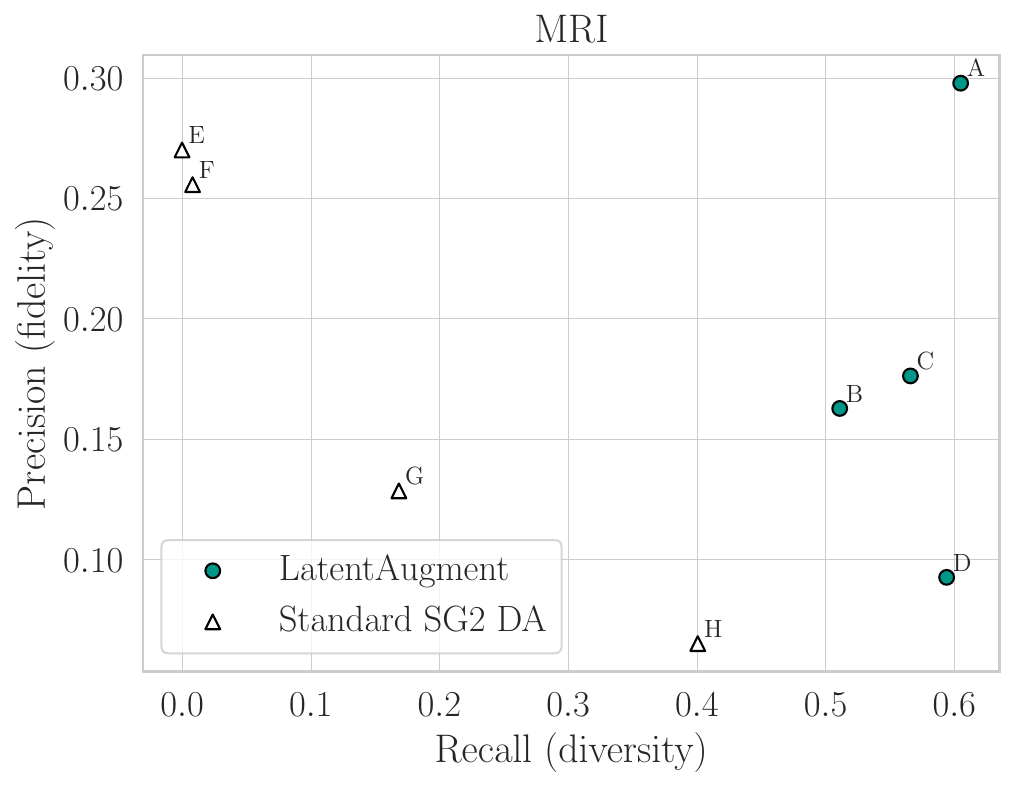}
         \label{fig:pr_mri}
     \end{subfigure}
    \centering
    \hfill
    \begin{subfigure}[b]{0.24\textwidth}
         \centering
         \includegraphics[width=\textwidth]{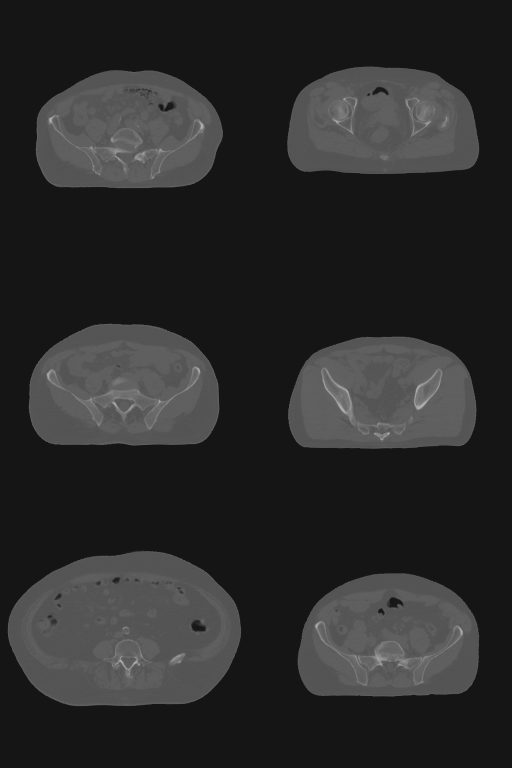}
         \caption{Real CT Images}
         \label{fig:real_ct}
    \end{subfigure}
    \begin{subfigure}[b]{0.24\textwidth}
         \centering
         \includegraphics[width=\textwidth]{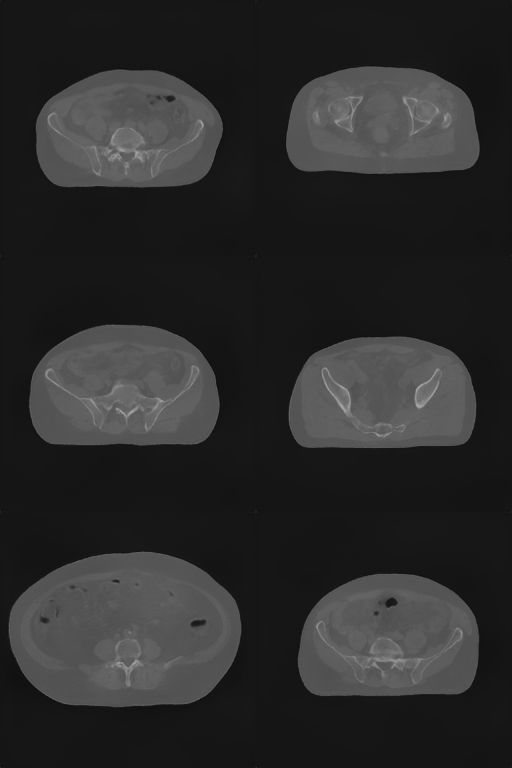}
         \caption{\textbf{A}}
         %(FID=$16.359$) }
         \label{fig:synthetic_ct_A}
    \end{subfigure}
    \begin{subfigure}[b]{0.24\textwidth}
         \centering
         \includegraphics[width=\textwidth]{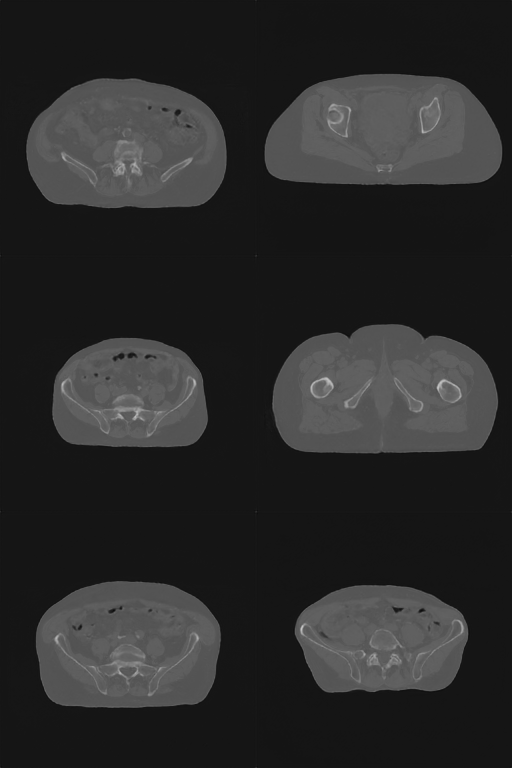}
         \caption{\textbf{H}}
         %(FID=$8.169$) }
         \label{fig:synthetic_ct_H}
     \end{subfigure}
    \begin{subfigure}[b]{0.24\textwidth}
         \centering
         \includegraphics[width=\textwidth]{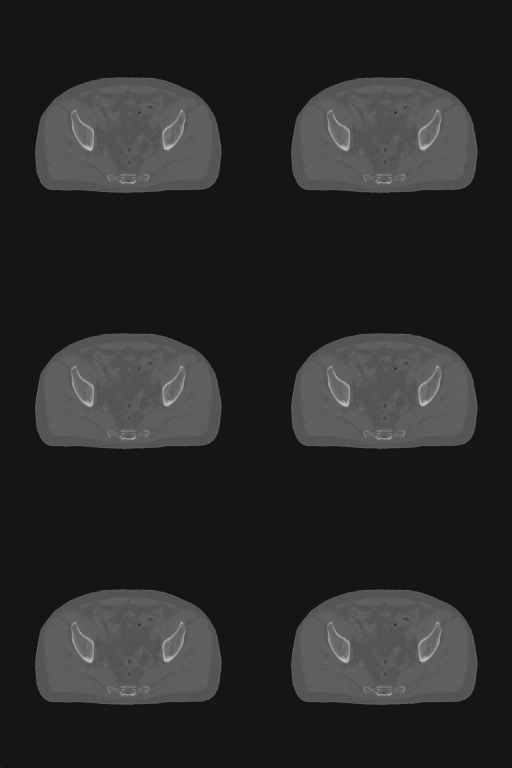}
         \caption{\textbf{E}}
         % (FID=$183.291$)}
         \label{fig:synthetic_ct_E}
     \end{subfigure}
         \hfill
    \begin{subfigure}[b]{0.24\textwidth}
         \centering
         \includegraphics[width=\textwidth]{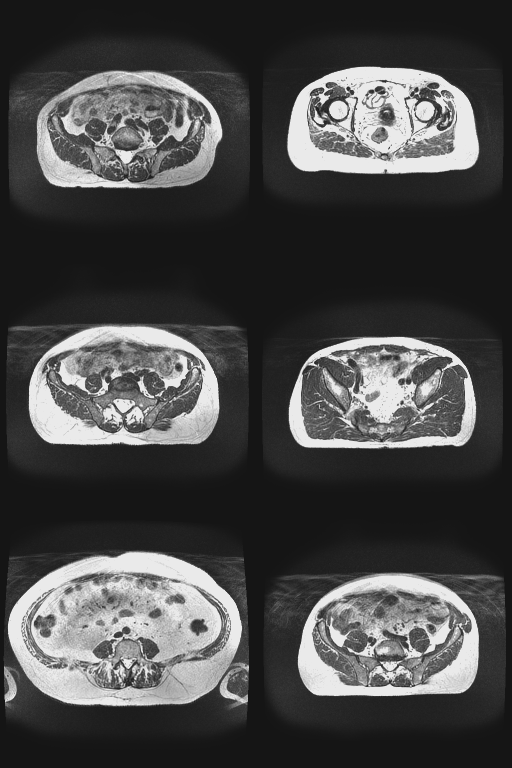}
         \caption{Real MRI Images}
         \label{fig:real_mri}
    \end{subfigure}
    \begin{subfigure}[b]{0.24\textwidth}
         \centering
         \includegraphics[width=\textwidth]{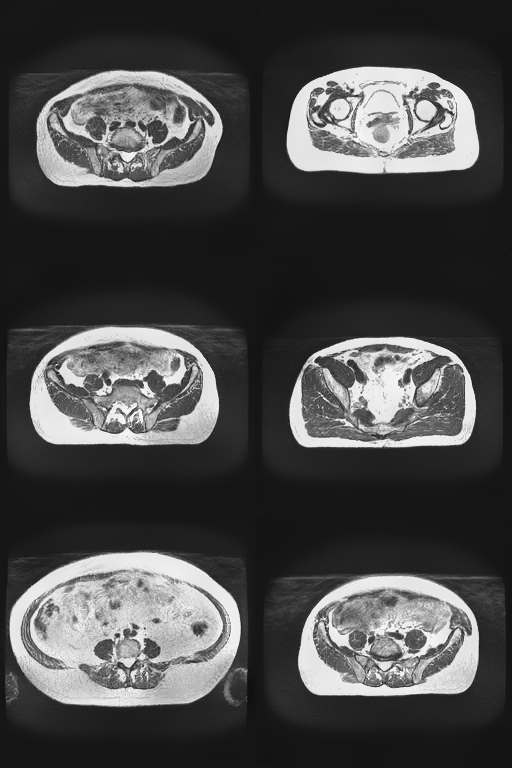}
         \caption{\textbf{A}}
         % (FID=$6.958$) }
         \label{fig:synthetic_mri_A}
    \end{subfigure}
    \begin{subfigure}[b]{0.24\textwidth}
         \centering
         \includegraphics[width=\textwidth]{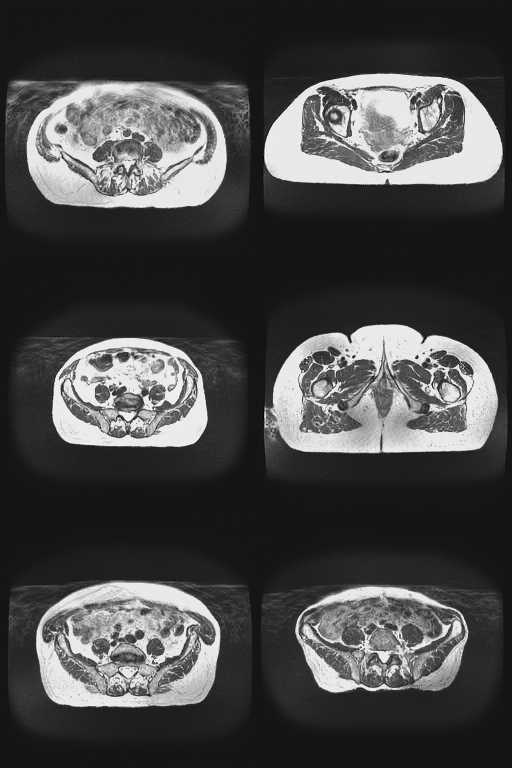}
         \caption{\textbf{H}}
         % (FID=$8.123$) }
         \label{fig:synthetic_mri_H}
     \end{subfigure}
    \begin{subfigure}[b]{0.24\textwidth}
         \centering
         \includegraphics[width=\textwidth]{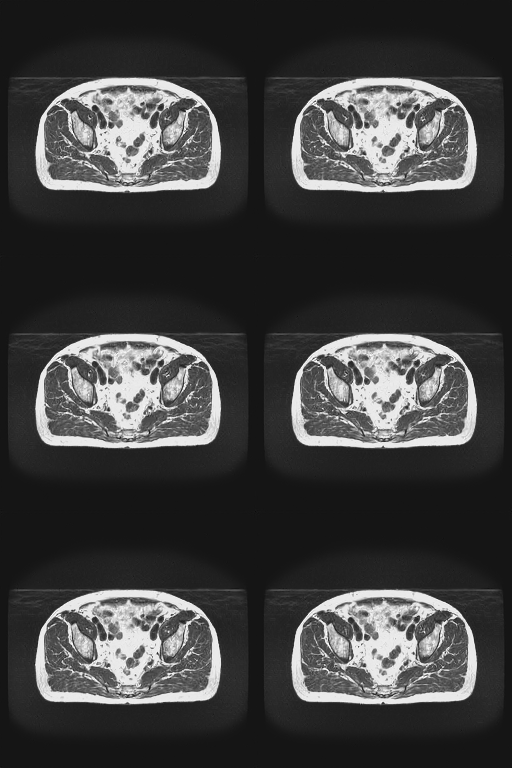}
         \caption{\textbf{E}}
         %(FID=$160.134$)}
         \label{fig:synthetic_mri_E}
     \end{subfigure}
     \caption{First row: precision-recall comparison of LatentAugment (green circle) and Standard SG2 DA (white triangles), respectively. 
     Second and third rows: qualitative examples of real and augmented samples  for configurations A, H, and E.}
     \label{fig:tackle_trilemma}
\end{figure*}

An effective DA method should satisfy the three key requirements mentioned in sections~\ref{sec:introduction} and~\ref{subsec:motivations}: high sample quality, diversity and mode coverage, and fast sampling~\cite{xiao2021tackling}.
In this respect, SG2 generates high-quality samples rapidly, as shown in \figureautorefname~\ref{fig:visual_inspection}, but it fails to guarantee mode coverage.

While in previous subsections we compared the performance of the proposed LatentAugment to the Standard SG2 DA, here we evaluate how LatentAugment and Standard SG2 samples compare in terms of diversity and fidelity.
We sampled 50,000 augmented images from each method and evaluated the precision and recall~\cite{kynkaanniemi2019improved} with respect to the real training set summarising the results in  \figureautorefname~\ref{fig:tackle_trilemma}.
The two uppermost plots show the precision-recall values achieved for both methods separately considering CT and MRI: to this end, we first generate the paired sCT-sMRI images, and then we compare the single mode to the real  CT and MRI training sets, respectively.
Note that to maximise mode coverage we seek the highest recall, while to maximise the visual quality of the image we seek the highest precision.

We tested eight configurations in total, four for each DA method.
For Standard SG2 DA we used the truncation trick~\cite{karras2019style} to investigate the diversity-fidelity trade-off:  a truncation of $1.0$ means searching for the maximum diversity, while a truncation of $0.0$ means searching for the maximum fidelity for the SG2 synthetic images according to the formula $w^{'} = \overline{w} + \psi (w - \overline{w})$, where $\psi \in [0, 1]$ and denotes the truncation parameter and $\overline{w}$ represents the average learned representation in the training data~\cite{karras2019style}.
We sampled four values of $\psi$, once for each letter reported in the plot, \ie, for E: 0.0, F: 0.3, G: 0.7, H: 1.0 (white triangles in the uppermost plots in \figureautorefname~\ref{fig:tackle_trilemma}). \bb
For LatentAugment, we randomly sampled four parameter configurations (A, B, C, and D in green in the same plots).
Observing such plots we notice that  LatentAugment, even randomly sampling its hyper-parameters, always beats Standard SG2 DA in terms of diversity since all the green triangles are further to the right on the diversity x-axis compared to white triangles, a finding worth for both modalities.
Moreover, turning now our attention to the fidelity y-axis while keeping on the  Standard SG2 DA  configuration with the best diversity (H), LatentAugment shows a comparable or higher fidelity.
This is not a limitation of LatentAugment because it is known that SG2 already generates high-quality samples~\cite{xiao2021tackling}. 
Thus, we conclude that LatentAugment can synthesise more diverse images than Standard DA SG2, having at least an equal visual quality.
 
The second and third rows of \figureautorefname~\ref{fig:tackle_trilemma}  offer a visual comparison of the images generated from both approaches compared to the real ones.
From left to right, we report six real CT (MRI) images, the augmentation results using configuration A for LatentAugment (best diversity and fidelity), and the augmentation results using configurations H and E for Standard SG2 DA, which correspond to maximum diversity and fidelity, respectively.
When observing the real and LatentAugment images in  Figures~\ref{fig:real_ct}, \ref{fig:real_mri} and Figures~\ref{fig:synthetic_ct_A}, \ref{fig:synthetic_mri_A}, respectively, we notice that our method infers in the real images new source of variance than traditional augmentation approaches.
Indeed, the LatentAugment images provide a smooth variation of the real ones while retaining the main content (main body structure) and style (texture, colour, etc.).
This allows LatentAugment to create realistic but diverse images with respect to those in the training set, avoiding samples that are out-of-distribution.
Furthermore, the comparison of Figures~\ref{fig:synthetic_ct_H}, \ref{fig:synthetic_mri_H}, \ref{fig:synthetic_ct_E}, \ref{fig:synthetic_mri_E} against real images shows that such a  relation is not satisfied by  Standard SG2 DA, which provides synthetic samples that are randomly sampled.

With reference to the third issue of the trilemma, which is related to the throughput of the generation, \tableautorefname~\ref{tab:performances}) shows that LatentAugment has a larger throughput than Standard SG2 DA. 
Nevertheless, this value is still reasonable compared to the inference time of recently emerged diffusion models~\cite{sohl2015deep,ho2020denoising,rombach2022high}.
Indeed, for the sake of comparison, we run a pre-trained latent diffusion model~\cite{rombach2022high} proposed to reduce the computational requirements compared to pixel-based diffusion models.
In the sampling procedure, we set the number of steps to create an image equal to 50.
It takes $16.8$ seconds to generate a single $256 \times 256$ image using a single NVIDIA A100 GPU, which is 2.4 times faster than the NVIDIA RTX A5000 GPU used in our experiments; we are forced to change GPU for memory issues. 
Thus, we deem that LatentAugment does not break the sampling speed requirement of the trilemma.
Moreover, the increased computational cost only applies when training the downstream model and not during inference.
Hence, once deployed, the downstream model will benefit from better performance thanks to LatentAugment, without any added computational cost.

\section{Conclusion} \label{sec:conclusion}

In this work we propose LatentAugment, a new method that navigates the latent space to improve the diversity and mode coverage of GAN synthetic images, enabling their adoption for DA purposes.
LatentAugment starts from the real image's latent representation and steers the latent space to maximise the spatial, perceptual, and semantic diversity of the generated images.
Moreover, it controls fidelity by maximising the realness score of the augmented images.

When compared to Standard DA and Standard SG2 DA, we demonstrated the feasibility of LatentAugment because it increased the generalisation performance of a deep model in the downstream application of MRI-to-CT translation.
Moreover, LatentAugment consistently improves GAN-generated images in terms of precision and recall, implying improved mode coverage while maintaining high-quality outputs, thus fulfilling the missing criteria of diversity and mode coverage in the generative learning trilemma.
A reflection on this work highlights three main avenues for future work.
The first concerns the increased computational overhead required for our policy to manipulate the real images in the GAN latent space.
In this respect, it would be warranted to investigate more computationally tractable approaches to steer the latent vectors in the GAN latent space for DA purposes.
As recently emerged diffusion models have shown promising results in image quality and mode coverage, the second direction of future investigation will compare our approach to such models. 
Third, let us remember that LatentAugment is now independent of the downstream task, which alleviates the need for domain expertise and makes the method work for all tasks. 
Nevertheless, we plan to develop a variation of LatentAugment to incorporate information from the downstream task, \eg, performance or overfitting issues, that will let us deepen how much this could be beneficial. 

\section*{Acknowledgements}

This research was partially supported by  Lion's Cancer Research Foundation in Northern Sweden (Grant No.~LP 18-2182 and No.~LP 22-2319). 
We also acknowledge financial support from PNRR MUR project PE0000013-FAIR (Italy). 
Resources provided by the National Academic Infrastructure for Supercomputing in Sweden (NAISS) and the Swedish National Infrastructure for Computing (SNIC) at Alvis @ C3SE partially funded by the Swedish Research Council through grant agreements no. 2022-06725 and no. 2018-05973.

\section*{Author contribution
}
L.T.: Conceptualization, Methodology, Software, Validation, Formal analysis, Investigation, Data Curation, Writing - Original Draft, Writing - Review \& Editing, Visualization. 
M.H.V.: Software, Visualization.
P.S.: Conceptualization, Methodology, Validation, Formal analysis, Writing - Review \& Editing, Supervision.
T.L.: Conceptualization, Methodology, Validation, Formal analysis, Writing - Review \& Editing, Supervision.
\bibliographystyle{ieeetr}

\balance
\bibliography{main.bib}

\end{document}